\documentclass[journal]{IEEEtran}
\usepackage{cite}
\usepackage{amsmath,amssymb,amsfonts}
\usepackage{algorithmic}
\usepackage{graphicx}
\usepackage{textcomp}
\usepackage{xcolor}
\usepackage{booktabs}
\usepackage{multirow}
\usepackage{hyperref}
\usepackage[font=footnotesize,labelfont=bf]{caption}
\usepackage{subcaption}
\usepackage{siunitx}
\usepackage{enumitem}
\usepackage[htt]{hyphenat}
\usepackage[section]{placeins}
\usepackage{xurl}

\graphicspath{{./}}

\begin{document}

\title{Achieving Near-Zero-Overhead Multi-Model Hierarchical Classification in Real-Time Detection Pipelines}

\author{Vaishnav~Raju\\
{\small Newspace Research and Technologies, Bengaluru, India}\\
{\small\texttt{\textbf{vaishnav.raju@newspace.co.in}}}}

\maketitle

\begin{abstract}
Edge-deployed vision systems in target recognition, surveillance, autonomous vehicles, and drone domains require hierarchical inference pipelines where a detection model identifies objects of interest and downstream classifiers provide fine-grained attribute analysis. Running all models on the GPU creates a serial bottleneck that limits real-time throughput as pipeline stages grow. Modern edge SoCs pair GPUs with dedicated neural accelerators (NPUs, DLAs) capable of concurrent execution, yet deploying custom models on these accelerators remains impractical due to strict operator constraints, quantization incompatibilities, and an undocumented end-to-end pipeline. We target NVIDIA Jetson DLA cores as the representative platform. We present a five-step methodology for zero GPU fallback DLA INT8 deployment of classification backbones, comprising architecture adaptation, manual dynamic range workaround to rescue TensorRT's implicit quantization (recovering 94.0\% accuracy from implicit quantization's 75\%) for rapid pipeline validation before explicit quantization, quantization-aware training, ONNX graph surgery for DLA compilation, and a concurrent GPU-detection/DLA-classification inference pipeline. We document nine engineering constraints with root-cause analysis and generalizable solutions. Validation on a dual-head person attribute classifier running on DLA alongside a GPU object detector on a Jetson Orin NX demonstrates near-zero pipeline overhead (12.5 vs.\ 13.3~FPS detector-only at 1080p), with dual-DLA scaling at no additional cost. The methodology is backbone-agnostic and generalizes to any detection-classification edge pipeline.

\end{abstract}


\section{Introduction}
\label{sec:introduction}

Edge-deployed computer vision systems increasingly require multi-stage inference pipelines. In advanced target recognition (ATR), surveillance, autonomous vehicles, and drone payloads, a detection model first identifies objects of interest, then one or more downstream classifiers provide fine-grained attribute analysis such as person type, pose, vehicle make or color, and presence of weapons or equipment. Each additional classification stage adds latency when all models share the GPU, creating a serial bottleneck that degrades real-time throughput as pipeline complexity grows.

NVIDIA Jetson SoCs provide a heterogeneous compute architecture comprising a GPU and one or two Deep Learning Accelerator (DLA) cores. On the Orin NX, this amounts to 100~TOPS of total compute: 60~TOPS on the GPU and $2 \times 20$~TOPS across two DLA cores~\cite{nvidia_orin_datasheet}. In typical deployments, the DLA cores sit entirely idle, representing 40\% of available hardware capacity going unused. Offloading classification to DLA enables concurrent execution where the GPU processes detections on the current frame while the DLA classifies crops from the previous frame, making classification effectively free in terms of pipeline latency. This benefit is detector-agnostic and grows with GPU model size, as heavier detectors (e.g., RT-DETR, transformer-based architectures) consume more GPU resources, leaving even less room for a concurrent GPU classifier.

However, deploying custom-trained models on DLA is an undocumented engineering challenge. DLA supports a restricted subset of operators, rejecting common layers such as \texttt{Flatten}, \texttt{nn.Linear} (which exports as \texttt{Gemm}/\texttt{MatMul}), \texttt{GlobalAveragePool} (which exports as \texttt{ReduceMean}), and \texttt{Concat} on certain tensor dimensions. TensorRT has deprecated implicit INT8 quantization in favor of explicit quantization via \texttt{QuantizeLinear}/\texttt{DequantizeLinear} (Q/DQ) nodes in the ONNX graph~\cite{nvidia_dla_docs}, yet DLA cannot execute these Q/DQ nodes directly so they must be stripped and converted to a calibration cache before engine compilation. Standard Post Training Quantization (PTQ) entropy calibration (\texttt{ENTROPY\_CALIBRATION\_2}) produces significantly degraded accuracy for DLA INT8 deployment: 75\% with ReLU6 and as low as 66\% per-head with standard ReLU, both well below the FP32 baseline. This represents a critical failure mode that is entirely undocumented in the literature.

To our knowledge, prior studies and vendor documentation do not provide a complete, experimentally validated methodology for zero-fallback deployment of custom INT8 quantization-aware-trained classifier models on DLA together with near-zero-overhead detection-classification pipeline integration. Prior studies on Jetson multi-accelerator inference focus on scheduling and contention analysis~\cite{tayal2025multiinstance, majeed2026scheduling} or layer-level parallelism within a single model~\cite{cpcnn2023}, but do not document the full end-to-end engineering process from architecture adaptation through quantization (both PTQ and QAT) to DLA compilation for custom classifiers.

\subsection{Contributions}

This paper makes the following contributions:

\begin{enumerate}[label=\textbf{\arabic*}., leftmargin=*, itemsep=2pt]
    \item A systematic, five-step methodology for adapting standard classification backbones (ResNet family) to achieve zero GPU fallback DLA deployment, including operator replacement, multi-head fusion, and explicit quantizer insertion.

    \item The discovery and root-cause analysis of PTQ entropy calibration failure on DLA INT8 where the accuracy degrades to 75\% with ReLU6 and further to 66--69\% with standard ReLU, 19--29 percentage points below the FP32 baseline. We show that structurally derived manual dynamic ranges recover accuracy to 94.0\%. ReLU6 additionally benefits manual range derivation, as its known output bounds $[0, 6]$ provide a structural anchor for INT8 ranges without calibration data.

    \item A GPU+DLA parallel execution architecture where detection runs on the GPU and classification runs on DLA via separate CUDA streams, with a frame $N{-}1$ design that makes classification near-zero-overhead.

    \item An extensible multi-head classification architecture with independent gradient control (\texttt{detach\_head\_b}), where additional attribute heads can be added without retraining existing ones.

    \item The complete DLA compilation pipeline: from a per-layer quantized backbone with \texttt{QuantConv2d} and skip-connection quantizers, through fused-head ONNX export with quantization-scale merging, to a model-specific graph surgery script that extracts per-tensor INT8 scales for all layers, including residual addition, propagates them through pass-through operators, removes Q/DQ nodes, and produces a TensorRT calibration cache for DLA engine build.

    \item Documentation of nine engineering constraints (C1-C9) encountered during development, each with failure mode, root cause, and generalizable solution.
\end{enumerate}

The remainder of this paper is organized as follows. Section~\ref{sec:related_work} reviews related work on edge deployment, quantization, and multi-task classification. Section~\ref{sec:methodology} presents the five-step DLA adaptation methodology with the constraints table. Section~\ref{sec:case_study} validates the methodology on a person attribute classification task. Section~\ref{sec:experiments} provides quantitative results including accuracy, latency, and ablation studies. Section~\ref{sec:discussion} discusses generalization, hardware utilization, and limitations. Section~\ref{sec:future_work} outlines extensions including structured sparsity and backbone scaling. Section~\ref{sec:conclusion} concludes.

\section{Related Work}
\label{sec:related_work}

\subsection{Edge Deployment and Quantization}

The NVIDIA Jetson platform is the dominant edge SoC for real-time vision, yet most deployments utilize only the GPU, leaving DLA cores idle. Prior work has characterized multi-instance DNN inference across GPU and DLA~\cite{tayal2025multiinstance} and surveyed scheduling on heterogeneous edge accelerators~\cite{majeed2026scheduling}, but neither addresses custom model adaptation for DLA or INT8 quantization. CP-CNN~\cite{cpcnn2023} parallelizes object detectors across GPU and DLA via \emph{intra-model} layer splitting, distributing layers of a single model across accelerators. Our work differs fundamentally by deploying \emph{separate models} on each accelerator as an inter-model parallel pipeline, with full INT8 quantization and documented DLA operator constraints. NVIDIA's own DLA tutorials~\cite{nvidia_jetson_dla_tutorial,nvidia_trt_work_with_dla} demonstrate toy workflows with FP16 fallback but do not provide a complete custom-model adaptation flow covering operator surgery, Q/DQ handling, calibration-cache generation, and zero-fallback validation.

PTQ and QAT are well-established for edge inference~\cite{gholami2021survey, jacob2018quantization, liang2021pruning, nagel2021whitepaper}, with advances in learned step sizes~\cite{esser2020lsq}, post-training 4-bit quantization~\cite{banner2019posttraining}, prune-quantize-distill pipelines~\cite{chae2025pruneqd}, and coreset-based QAT~\cite{huang2024coreset}. These target GPU inference exclusively. DLA introduces a distinct quantization challenge that GPU-oriented methods do not encounter. TensorRT's transition from implicit to explicit quantization requires Q/DQ nodes in the ONNX graph, yet DLA cannot execute these nodes natively. They must be stripped via custom graph surgery and their scale factors extracted into a calibration cache, a pipeline not documented in any prior work. Our work addresses this gap and additionally documents the failure of TRT's implicit INT8 with entropy calibration on DLA, which degrades accuracy by 19--29 percentage points depending on activation function. This failure mode has not been reported in prior quantization literature, as existing studies evaluate calibration exclusively on GPU backends where entropy calibration performs reliably.

\subsection{Detection-Classification Pipelines}

Two-stage detection-classification pipelines are standard in ATR~\cite{safdar2023yoloatr}, equipment detection~\cite{wang2023weapon}, and person attribute recognition~\cite{wang2022par,bekele2019arl}. YOLO deployment on edge devices~\cite{Rey_2025} and multi-attribute classification have been studied extensively, but all deploy on a single GPU, creating a serial bottleneck as pipeline stages grow. Adding a classifier to a GPU-only pipeline serializes inference and reduces detector throughput proportionally to classifier latency. Our work eliminates this bottleneck by offloading classification to DLA using a frame $N{-}1$ asynchronous design, where the DLA processes crops from the previous frame while the GPU detects on the current frame, achieving near-zero overhead on the detection path. The multi-head architecture with independent Conv$1\times1$ heads and head fusion into a single Conv2d for ONNX export is specifically designed for DLA constraints, where Concat operations on certain tensor dimensions cause compilation failures. This enables adding classification heads with zero marginal DLA cost, a property that GPU-only pipelines cannot offer.

\section{Methodology: DLA-Ready Classifier Adaptation}
\label{sec:methodology}

This section presents a general, five-step methodology for adapting classification networks to run entirely on NVIDIA DLA with zero GPU fallback. The methodology is both backbone-agnostic and detector-agnostic, as all steps operate on standard DLA-supported operators rather than architecture-specific constructs, and the pipeline integration works with any GPU-based detector. While we illustrate each step with our ResNet-34 case study, it applies directly to any residual architecture (ResNet-18/34/50) and generalizes to other backbone families (MobileNetV2/V3, EfficientNet-Lite) subject to per-operator DLA verification on the target Jetson device JetPack version.

\subsection{Engineering Constraints Encountered}
\label{sec:constraints}

Before presenting the methodology, we document the nine engineering constraints discovered during development. Each row in Table~\ref{tab:constraints} represents a failure encountered, its root cause, and the solution that became a step in the final methodology. The ablation study (Section~\ref{sec:ablation}) provides structured evidence that each fix was necessary.

\begin{table*}[t]
\centering
\caption{Engineering constraints encountered during DLA deployment. Each failure and its solution became a step in the methodology.}
\label{tab:constraints}
\small
\begin{tabular}{@{}clp{3.8cm}p{4.2cm}p{4.5cm}@{}}
\toprule
\textbf{\#} & \textbf{Step} & \textbf{What Failed} & \textbf{Root Cause} & \textbf{Solution} \\
\midrule
\phantomsection\label{constr:c1}C1 & 1 & \texttt{nn.Linear} head rejected by DLA & Exports as \texttt{Flatten}+\texttt{MatMul}/\texttt{Gemm}, none are DLA-supported ops & Replace with \texttt{Conv2d(512$\to$1, k=1)}. It is mathematically identical, exports as DLA-native \texttt{Conv} \\
\addlinespace
\phantomsection\label{constr:c2}C2 & 1 & \texttt{AdaptiveAvgPool2d(1)} rejected & Exports as \texttt{ReduceMean} and is not supported by DLA, but supports \texttt{AveragePool} with explicit kernel size ($\leq$8) only & Replace with \texttt{AvgPool2d(k=7)} for $224\times224$ input \\
\addlinespace
\phantomsection\label{constr:c3}C3 & 1 & \texttt{torch.cat} on two $[N,1,1,1]$ heads crashes DLA compiler & \texttt{Concat} on 1$\times$1 spatial tensors triggers dimension assertion by the DLA TensorRT compiler and fails. & At export, fuse heads into a single \texttt{Conv2d(512$\to$2, k=1)} by stacking individual head weights along the output-channel dimension \\
\addlinespace
\phantomsection\label{constr:c4}C4 & 2 & TRT implicit INT8 quantization (\texttt{trtexec --int8}) deprecated & TRT 10.x requires explicit quantization for DLA INT8 & Custom backbone with \texttt{QuantConv2d} at every conv + \texttt{skip\_connection\_quantizer} at every skip connection \\
\addlinespace
\phantomsection\label{constr:c5}C5 & 2 & AvgPool and Sigmoid run in FP16 (can fallback to GPU) & \texttt{QuantConv2d} only covers conv boundaries, so non-conv ops have no Q/DQ & Add explicit \texttt{TensorQuantizer} for \texttt{post\_backbone} (before AvgPool) and \texttt{post\_head} (before Sigmoid) \\
\addlinespace
\phantomsection\label{constr:c6}C6 & 2 & \texttt{post\_output\_quantizer} after Sigmoid causes degenerate amax & Sigmoid output $\in [0,1]$ is nearly uniform. Entropy calibration cannot find meaningful threshold & Removed. Sigmoid on $[16,2,1,1]$ is 32 ops, INT8 vs FP16 makes no measurable difference \\
\addlinespace
\phantomsection\label{constr:c7}C7 & 3 & PTQ entropy calibration at 75\% accuracy for ReLU6, 19--20pp below FP32 baseline & Entropy selects degenerate clipping thresholds, and standard ReLU fares worse at 66--69\% per-head (A4) & Structurally derived manual ranges: input $[-4,4]$, intermediate $[-8,8]$, output $[-1,1]$. Recovers accuracy to 94.0\% \\
\addlinespace
\phantomsection\label{constr:c8}C8 & 4 & QAT ONNX with Q/DQ nodes fails DLA compilation & DLA expects QAT weights with a calibration cache generated from the stripped Q/DQ nodes & A custom graph surgery script traverses the ONNX graph, extracts per-tensor INT8 scales from Q/DQ node pairs, propagates through pass-through ops and residual connections, strips Q/DQ nodes, and produces a TRT calibration cache \\
\addlinespace
\phantomsection\label{constr:c9}C9 & 4 & The custom graph surgery script fails on raw PyTorch ONNX & No-op \texttt{Cast}/\texttt{Identity} nodes break graph traversal for scale propagation & Run \texttt{onnxsim.simplify()} after ONNX export, before running the graph surgery script \\
\bottomrule
\end{tabular}
\end{table*}

\subsection{Step 1: DLA-Safe Architecture Adaptation}
\label{sec:step1}

DLA supports a restricted operator set. The first step audits the target architecture and replaces unsupported operators with DLA-native equivalents, preserving mathematical equivalence.

\textbf{Operator replacements.} Three standard classification layers must be replaced (constraints C1-C3):

\begin{itemize}[itemsep=2pt]
    \item \texttt{nn.Linear(C, num\_classes)} $\to$ \texttt{Conv2d(C, num\_classes, kernel\_size=1)}. When the spatial dimension is $1\times1$ (after global pooling), a $1\times1$ convolution is mathematically identical to a linear layer but exports as a DLA-native \texttt{Conv} operation rather than \texttt{Gemm}/\texttt{MatMul}.

    \item \texttt{AdaptiveAvgPool2d(1)} $\to$ \texttt{AvgPool2d(kernel\_size=k)} where $k$ equals the spatial dimension of the final feature map. DLA supports \texttt{AveragePool} with explicit kernel size (maximum window size of 8) but not the adaptive variant, which exports as \texttt{ReduceMean}.

    \item Multi-head output concatenation $\to$ fused single-head convolution. When multiple classification heads each produce $[N,1,1,1]$ tensors, \texttt{torch.cat} exports as an ONNX \texttt{Concat} node on $1\times1$ spatial dimensions. The \texttt{trtexec} DLA compiler crashes on this configuration with a dimension assertion failure, suggesting a fallback to the GPU. The solution is to fuse $K$ heads of \texttt{Conv2d(C$\to$1)} into a single \texttt{Conv2d(C$\to$K)} at export time. Concretely, each head has a weight tensor of shape $[1, C, 1, 1]$. At export, these are concatenated along the output-channel dimension- \texttt{torch.cat([head\_0.weight, head\_1.weight, \ldots], dim=0)} to form a single $[K, C, 1, 1]$ kernel, with biases similarly stacked into $[K]$. The resulting single \texttt{Conv2d(C$\to$K)} produces all $K$ outputs in one convolution pass with no \texttt{Concat} node in the exported graph. This is strictly equivalent to running each head independently, since $1\times1$ convolutions on the same input are separable by output channel. DLA executes it as a single native op, eliminating the compiler crash and reducing kernel launch overhead.
\end{itemize}

\textbf{Activation function.} We recommend replacing the standard ReLU activations in ResNet-34 with ReLU6 ($\min(\max(x, 0), 6)$). This is not a strict requirement. Our ablation (Section~\ref{sec:ablation}, A4) confirms that standard ReLU also achieves competitive accuracy (94.0\% vs.\ 95.0\%), and the full methodology applies to either activation. ReLU6 is the recommended choice because it provides strictly better accuracy and AUC at every operating point (QAT, PTQ manual, PTQ entropy) with no latency cost, and its bounded output range $[0, 6]$ enables structurally derived dynamic ranges without calibration data (Step~3). For backbones with standard ReLU, PTQ manual ranges remain applicable using activation bounds derived from empirical maximum values, and QAT with percentile calibration produces well-calibrated INT8 scales without requiring bounded activations.

\textbf{Tensor format.} All tensors remain in 4D format ($N \times C \times H \times W$) throughout the network. No \texttt{Flatten} or \texttt{Reshape} operations are used. The combination of explicit \texttt{AvgPool2d} (reducing spatial dimensions to $1\times1$) and \texttt{Conv2d} heads (operating on $1\times1$ feature maps) keeps the graph entirely in 4D.

\subsection{Step 2: Explicit Quantization with Per-Layer Q/DQ Coverage}
\label{sec:step2}

TensorRT 10.x has deprecated implicit INT8 quantization~\cite{nvidia_dla_docs}. INT8 engines now require explicit \texttt{QuantizeLinear}/\texttt{DequantizeLinear} (Q/DQ) nodes in the ONNX graph, specifying per-tensor quantization scales. This step inserts Q/DQ nodes to achieve complete INT8 coverage with zero gaps.

\textbf{Quantized backbone (C4).} We modify the standard ResNet-34 backbone with the help of TensorRT's \texttt{pytorch-quantization}~\cite{nvidia_pytorchquant} library. Every \texttt{Conv2d} is replaced with \texttt{QuantConv2d}, which wraps each convolution with input and weight \texttt{TensorQuantizer} modules that insert Q/DQ nodes in the model's graph. Critically, each \texttt{BasicBlock} includes a \texttt{skip\_connection\_quantizer}, which is a \texttt{TensorQuantizer} on the skip-connection path before the element-wise addition. Without this, the residual add operation lacks INT8 scale information, and the DLA compiler falls back to FP16 for all residual blocks (C4).

\textbf{Explicit quantizers at non-conv boundaries (C5).} \texttt{QuantConv2d} only inserts Q/DQ nodes at the convolution input/weight/output boundaries. Tensor boundaries that do not pass through a convolution, specifically the gap between the last convolutional layer's output and \texttt{AvgPool2d}, and the gap between the head convolution's output and \texttt{Sigmoid}, have no Q/DQ coverage. Without explicit quantizers at these boundaries, the graph surgery script finds no INT8 scale for those tensors and assigns FP16 precision to everything downstream, causing GPU fallback.

We add two explicit \texttt{TensorQuantizer} modules:
\begin{itemize}[itemsep=2pt]
    \item \texttt{post\_backbone\_quantizer}: placed after \texttt{layer4} (before \texttt{AvgPool2d}). \texttt{AvgPool2d} is a pass-through operator for scale propagation, so the graph surgery script propagates the INT8 scale through it to the head input. No additional quantizer is needed between \texttt{AvgPool2d} and the head convolution.
    \item \texttt{post\_head\_quantizer}: placed after each head's convolution output (before \texttt{Sigmoid})
\end{itemize}

\textbf{Why post-Sigmoid quantization was removed (C6).} An initial design included a \texttt{post\_output\_quantizer} after the Sigmoid activation. During calibration, this quantizer produced degenerate amax values (scipy division-by-zero) as the Sigmoid output $\in [0,1]$ is a nearly uniform distribution where entropy calibration cannot find a meaningful clipping threshold. Since Sigmoid operates on only 32 elements ($[16, 2, 1, 1]$ for batch-16 with 2 heads), the INT8 vs.\ FP16 difference is immeasurable. We removed this quantizer entirely.

\subsection{Step 3: INT8 Quantization Strategy}
\label{sec:step3}

TensorRT 10.x has deprecated implicit INT8 quantization in favor of explicit quantization via Q/DQ nodes~\cite{nvidia_pytorchquant}. Following this recommendation, the primary production workflow uses QAT with explicit Q/DQ coverage (Section~\ref{sec:qat}). During development, however, we first attempted PTQ with implicit entropy calibration to establish a baseline. This revealed a fundamental failure mode in terms of accuracy and led to the discovery that structurally derived manual dynamic ranges, applied directly to the PTQ model (ReLU6 version) without any QAT training, recover accuracy to 94.0\%. This emerged as a useful alternative as it bypasses QAT entirely, requires only minutes of engineering work, and serves as a rapid deployment and testing path, useful for validating DLA operator compatibility and latency before committing QAT implementation and training time.

\subsubsection{PTQ with Manual Dynamic Ranges (C7)}
\label{sec:manual_ptq}

As an initial experiment, we applied the standard TRT implicit PTQ entropy calibration (\texttt{ENTROPY\_CALIBRATION\_2}) before committing to QAT training. This produced only 75\% accuracy, 19--20 percentage points below the FP32 baseline of 94.5\%.

\textbf{Root cause.} Entropy calibration is unreliable for DLA INT8 deployment regardless of activation function. We initially trained with standard ReLU, which produced even worse results (66--69\% per-head accuracy, Section~\ref{sec:ablation}, A4). Switching to ReLU6 improved entropy calibration to 75\%, but this is still 19--20 percentage points below the FP32 baseline. The underlying mechanism is that entropy minimization selects a small clipping threshold $T$ to minimize KL divergence between the FP32 and quantized distributions. Activations exceeding $T$ saturate at INT8 value 127, and this saturation propagates through 34 layers of residual blocks. Unbounded ReLU activations produce a wider histogram that causes entropy to select an even more aggressive threshold, explaining the worse performance with standard ReLU.

\textbf{Solution.} The entropy failure prompted us to examine the actual activation bounds analytically rather than relying on the calibration framework. We derive INT8 dynamic ranges structurally from the network's known activation constraints:

\begin{itemize}[itemsep=2pt]
    \item \textbf{Input tensors}: $[-4, 4]$ — ImageNet-normalized pixels have per-channel statistics within approximately $[-2.5, 2.5]$, so $[-4, 4]$ provides conservative headroom without distorting INT8 resolution.
    \item \textbf{Intermediate tensors}: $[-8, 8]$ — ReLU6 bounds post-activation feature maps to $[0, 6]$. Pre-activation values (conv outputs before ReLU6) and residual additions can exceed this, and $[-8, 8]$ covers the observed pre-activation range with headroom for residual accumulation across depth.
    \item \textbf{Output tensors}: $[-1, 1]$ — The \texttt{post\_head\_quantizer} sits after the classification head convolution but \emph{before} Sigmoid. These are logit values whose magnitude is bounded by the trained weight norms, and $[-1, 1]$ covers the logit range for a well-calibrated binary classifier where confident predictions produce logits of $\pm1$-$\pm2$.
\end{itemize}

These ranges are set directly on TensorRT network tensors via the \texttt{set\_dynamic\_range()} API, bypassing the calibration framework entirely. This approach recovers accuracy to 94.0\% with ROC-AUC $>0.97$ on both classification heads within 1\,pp of QAT while requiring no training time and significantly lower pipeline complexity. It is well-suited for rapid deployment and testing as it helps in validating DLA operator compatibility, measuring latency, and confirming end-to-end correctness before committing to QAT implementation and fine-tuning.

\textbf{Generalization:} Entropy calibration is unreliable for DLA INT8 deployment regardless of activation function (Section~\ref{sec:ablation}, A4 confirms even worse degradation with standard ReLU). Structurally derived manual dynamic ranges should be the first quantization strategy attempted. We found that bounded activations (ReLU6) additionally simplify manual range derivation, as the known output range provides a structural anchor without calibration data.

\subsubsection{Quantization-Aware Training}
\label{sec:qat}

QAT is the primary production workflow, chosen because TRT recommends explicit Q/DQ quantization for DLA INT8~\cite{nvidia_pytorchquant}. QAT fine-tunes the model with simulated quantization noise (fake-quantization), adapting weights to INT8 constraints while preserving explicit Q/DQ nodes for DLA engine compilation.

\textbf{Calibration:} Before fine-tuning, histogram collectors gather activation distributions across 32 or more calibration batches. The collected histograms are then processed to compute per-tensor amax (clipping threshold) values.

\textbf{Configurable calibration methods:} We implement three amax computation methods as a configurable parameter for systematic comparison:

\begin{itemize}[itemsep=2pt]
    \item \textbf{Entropy}: minimizes KL divergence between FP32 and INT8 distributions. Aggressive clipping produces suboptimal thresholds for DLA INT8 (Section~\ref{sec:manual_ptq}).
    \item \textbf{Percentile (99.99th)}: clips at the 99.99th percentile. Preserves the dense activation range while discarding transient residual-add spikes. Best for this model class.
    \item \textbf{MSE}: minimizes mean squared error. Sensitive to outliers from residual-add spikes.
\end{itemize}

Percentile calibration at the 99.99th percentile produces the best results for this model class as the 0.01\% of clipped values are transient residual-add spikes that carry no classification signal.

\textbf{Fused-head export.} For multi-head models, ONNX export requires merging per-head quantization parameters into the fused \texttt{Conv2d(C$\to$K)} kernel. The input amax is taken as the maximum across heads (conservative clipping), and per-channel weight amax values are concatenated along the output channel dimension.

\subsection{Step 4: ONNX Export and DLA Compilation}
\label{sec:step4}

\textbf{ONNX simplification (C9):} PyTorch's ONNX exporter produces a raw graph that contains no-op \texttt{Cast} nodes (inserted for dtype safety), \texttt{Identity} nodes (from residual paths), and redundant constant subgraphs. Although \texttt{torch.onnx.export()} performs constant folding by default, the resulting graph still retains \texttt{Cast} and \texttt{Identity} nodes that break the graph surgery script's scale-propagation traversal, which expects direct tensor-to-tensor connectivity. We apply \texttt{onnxsim.simplify()} after export to remove these remaining no-op nodes through graph-level dead-code elimination and constant propagation. The result is a clean graph where every tensor flows directly between computational operators, enabling reliable Q/DQ scale extraction in the next stage.

\textbf{Q/DQ stripping via custom graph surgery script (C8):} While TensorRT on GPU handles \texttt{QuantizeLinear}/\texttt{DequantizeLinear} (Q/DQ) nodes natively, DLA does not. It expects QAT weights and a calibration cache with per-tensor INT8 scales provided at engine build time, not inline Q/DQ nodes at runtime. The \texttt{graph\_surgery script} traverses the ONNX graph, extracts per-tensor scales from every Q/DQ node pair, propagates scales through pass-through operators (ReLU6, AveragePool, etc.), strips all Q/DQ nodes, and produces a clean ONNX model together with a TensorRT calibration cache consumable at DLA engine build time.

\textbf{DLA engine build:} The TensorRT engine is built via the Python API with the following configuration: INT8 precision enabled, DLA core assignment (\texttt{DLA\_CORE=0} or \texttt{1}), GPU fallback flag enabled (though zero layers will fall back), per-tensor dynamic ranges set from the calibration cache, and both optimization and calibration profiles configured. The build log is checked to verify zero GPU fallback layers.

\subsection{Step 5: Asynchronous Parallel Pipeline Integration}
\label{sec:step5}

With the classifier deployed on DLA, the final step integrates it into an asynchronous parallel inference pipeline where GPU detection and DLA classification execute concurrently on separate hardware with frame-level pipelining.

\subsubsection{CUDA Stream Architecture}

Each accelerator operates on a dedicated CUDA stream: the GPU detection model executes on the default stream, while each DLA engine is bound to its own stream via \texttt{torch.cuda.Stream()}. DLA core assignment is set at TensorRT runtime creation (\texttt{runtime.DLA\_core = 0} or \texttt{1}), binding each engine to a specific DLA core. Streams execute independently; work enqueued on the DLA stream does not block or contend with GPU stream execution.

\subsubsection{Frame \texorpdfstring{$N{-}1$}{N-1} Async Dispatch}

The pipeline processes each video frame in three phases:

\begin{enumerate}[itemsep=2pt]
    \item \textbf{Collect previous results:} At the start of frame $N$, call \texttt{stream.synchronize()} on the DLA stream to retrieve classification results from frame $N{-}1$. Since DLA inference ($\approx$20\,ms) completes well within a typical detector frame interval (e.g., $\approx$76\,ms for YOLOv11m at 1080p), this synchronization typically returns immediately with no stall. The margin grows further with heavier detectors such as RT-DETR, RF-DETR, RTMDet or other transformer-based architectures that have longer frame intervals.

    \item \textbf{GPU detection + crop preprocessing:} Run the detector on frame $N$ (GPU default stream). After detection, extract person crops on the GPU where each crop is sliced from the GPU-resident frame tensor, letterboxed to maintain aspect ratio, resized to $224\times224$ via bicubic interpolation, and ImageNet-normalized as in-place GPU operations. The preprocessed batch is a $[\text{batch}, 3, 224, 224]$ float32 tensor already resident in GPU memory. This crop preprocessing is the source of the measured 6--9\% pipeline overhead for an engine batch size of 16. If you reduce the batch size the preprocessing overhead will come down.

    \item \textbf{Non-blocking DLA enqueue:} On the DLA stream, copy the preprocessed crop tensor to the DLA engine's input buffer, then call \texttt{execute\_async\_v3(stream\_handle=...)} to enqueue DLA inference. This returns immediately and the CPU proceeds to frame $N{+}1$ while DLA executes concurrently. Device-to-host output copy is also enqueued asynchronously on the same stream.
\end{enumerate}

\begin{figure}[t]
    \centering
    \includegraphics[width=\columnwidth]{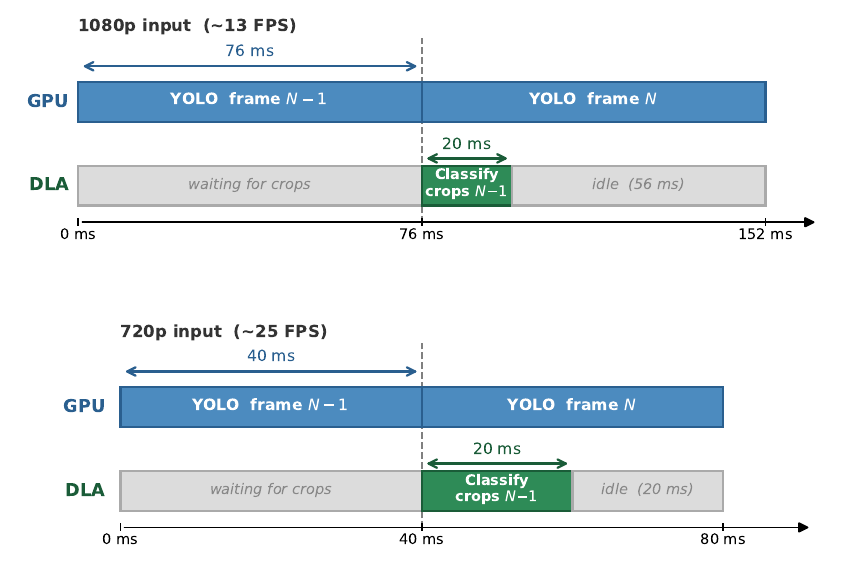}
    \caption{Frame $N{-}1$ parallel pipeline. The GPU runs detection on the current frame while DLA concurrently classifies crops extracted from the previous frame. The detector is application-specific (YOLOv11m in our experiments, but any GPU-based detector applies). Values are approximate. See Table~\ref{tab:pipeline} for measured throughput.}
    \label{fig:timing}
\end{figure}

The key property is that phases 2 and 3 overlap with the \emph{next} frame's phase 1. While the GPU processes frame $N$, DLA is classifying frame $N{-}1$'s crops. Classification results lag detection by one frame, which is acceptable at real-time video rates as objects rarely change attributes between consecutive frames. Fig.~\ref{fig:timing} illustrates this concurrent execution.

\subsubsection{Batch Padding}

DLA engines require a fixed batch size (16 in our configuration). When the number of detections is fewer than 16, the input tensor is zero-padded to the static batch size before enqueue. After synchronization, only the first $n$ outputs (corresponding to actual detections) are retained, and padded outputs are discarded.

\subsubsection{Multi-DLA Scaling}

With two DLA cores, a second classifier engine is instantiated on DLA core~1 with its own dedicated CUDA stream. Crop preprocessing occurs once on the first DLA stream, and a CUDA event is recorded after preprocessing completes. The second DLA stream waits on this event before reading the shared input tensor. Both DLA engines then execute concurrently and independently within the detector frame interval, producing results for different classification tasks on the same set of crops. Section~\ref{sec:experiments} confirms that dual-DLA throughput is identical to single-DLA, validating true hardware-level independence.

\section{Case Study: Person Attribute Classification}
\label{sec:case_study}

We validate the methodology on a dual-head person attribute classifier deployed alongside a YOLOv11m~\cite{yolo11_ultralytics} detector on an NVIDIA Jetson Orin NX. The pipeline operates on 1080p inputs because high-altitude aerial and elevated-viewpoint imagery produces small person targets that occupy a limited number of pixels. Consequently, high input resolution is necessary to preserve sufficient spatial detail for the detector to localize targets reliably and for the downstream classifier to distinguish fine-grained attributes from small crops. This section describes the specific instantiation, while the methodology (Section~\ref{sec:methodology}) is general.\footnote{YOLOv11m is used for experimental evaluation under its research license. Deployment configurations may substitute an appropriately licensed detector and the DLA classification methodology is detector-agnostic.}

\subsection{Task and Dataset}

The classifier performs two independent binary classification tasks on person crops extracted from YOLO detections, with two mutually independent attribute labels per image:

\begin{itemize}[itemsep=2pt]
    \item \textbf{Head A (Attribute~1)}: A binary person attribute based on visual appearance.
    \item \textbf{Head B (Attribute~2)}: A second independent binary person attribute.
\end{itemize}

Attribute names are anonymized as this work is part of a private R\&D program. The methodology generalizes to any binary person attributes, e.g., clothing type or color, pose category, or carried-object presence, and extends to vehicle attributes such as color, make, and type.

The two binary heads yield four combined attribute states, though the model is trained and evaluated on each head independently. Training uses an internal proprietary dataset of labeled person crops sourced from operational imagery and diverse external modalities, split 80\% / 10\% / 10\% into training, validation, and test subsets. The test set contains approximately 2,000 samples. Crops vary in resolution and framing, ranging from full-person to partial-torso views.

The dataset exhibits moderate class imbalance, addressed via the techniques described below.

\subsection{Class Imbalance Handling}
\label{sec:imbalance}

With two independent binary heads producing four joint attribute combinations, class imbalance is common in real-world datasets where certain combinations occur less frequently than others. For our dataset we address this issue via the per-head \texttt{pos\_weight} in the binary cross-entropy loss where Head~A $w^+=1.541$ and Head~B $w^+=1.641$. For sampling-level balancing we employ \texttt{WeightedRandomSampler} with a \texttt{min\_count} floor to cap the oversample ratio at $\text{max\_class} / \text{min\_count}$, ensuring rare combinations are seen frequently during training without being oversampled to the point of memorization.

\subsection{Model Architecture}

The Person Classification Network (PCN) instantiates the methodology from Section~\ref{sec:methodology}:

\begin{itemize}[itemsep=2pt]
    \item \textbf{Backbone}: A custom quantized ResNet-34~\cite{he2016resnet} with \texttt{QuantConv2d} at every convolutional layer, \texttt{skip\_connection\_quantizer} at every skip connection, and ReLU6 activations throughout, initialized from standard ImageNet-pretrained ResNet-34 weights.

    \item \textbf{Pooling}: \texttt{AvgPool2d(kernel\_size=7)} replacing \texttt{AdaptiveAvgPool2d(1)}.

    \item \textbf{Heads}: Two independent \texttt{Conv2d(512$\to$1, kernel\_size=1)} followed by \texttt{Sigmoid}. At ONNX export, both heads are fused into a single \texttt{Conv2d(512$\to$2)} by stacking each head's weight tensor along the output-channel dimension, eliminating the \texttt{Concat} node that causes DLA compiler failures (C3).

    \item \textbf{Explicit Quantizers}: \texttt{post\_backbone\_quantizer} (after \texttt{layer4}, before \texttt{AvgPool2d}) and \texttt{post\_head\_quantizer} (after head conv, before Sigmoid).

    \item \textbf{Optional Gradient Control}: \texttt{detach\_head\_b=True} stops gradients from Head~B propagating into the backbone when one attribute class has insufficient data, preventing gradient entanglement between heads.
\end{itemize}

\begin{table}[h]
\centering
\caption{FP32 training hyperparameters.}
\label{tab:fp32_training}
\begin{tabular}{@{}ll@{}}
\toprule
\textbf{Parameter} & \textbf{Value} \\
\midrule
Optimizer & AdamW \\
Epochs & 60 \\
Backbone learning rate & $1 \times 10^{-4}$ \\
Head learning rate & $5 \times 10^{-4}$ \\
Scheduler & Cosine annealing ($t_{\text{initial}}{=}60$, warmup$_t{=}3$) \\
Label smoothing & $\epsilon = 0.04$: $y_{\text{smooth}} = (1{-}\epsilon)\,y + \epsilon/K$ \\
\bottomrule
\end{tabular}
\end{table}

\begin{table}[h]
\centering
\caption{QAT training hyperparameters.}
\label{tab:qat_training}
\begin{tabular}{@{}ll@{}}
\toprule
\textbf{Parameter} & \textbf{Value} \\
\midrule
Epochs & 25 \\
Calibration batches & 32 \\
Calibration method & Percentile (99.99th) \\
Fine-tuning LR schedule & Same as FP32 \\
\bottomrule
\end{tabular}
\end{table}

\subsection{Training Protocol}

\subsubsection{FP32 Baseline}

Differential learning rates are critical. A uniform learning rate ($1 \times 10^{-3}$) across the pretrained backbone and randomly-initialized heads caused validation loss oscillations. Reducing the backbone rate to $1 \times 10^{-4}$ while keeping heads at $5 \times 10^{-4}$ resolved this as the backbone adapts gently without overwriting pretrained features, while heads converge aggressively from random initialization.

We performed multiple FP32 training runs and selected the checkpoint with the best validation metrics as the starting point for QAT fine-tuning.

Label smoothing with $\epsilon = 0.04$ applies the standard formulation $y_{\text{smooth}} = (1 - \epsilon)\,y + \epsilon/K$, yielding smoothed targets of 0.98 and 0.02 for $K{=}2$ classes. This prevents overconfident sigmoid outputs and improves binary cross-entropy stability by keeping targets away from 0 and 1.

\subsubsection{Quantization-Aware Training}

QAT starts from the FP32 checkpoint with quantizers enabled:

The 99.99th percentile calibration method was selected after systematic comparison (Section~\ref{sec:calibration_comparison}). It preserves high resolution for the dense $[0, 6]$ ReLU6 activation range while clipping the sparse tail from residual additions which contains transient values that carry no classification signal.

\section{Experiments}
\label{sec:experiments}

\subsection{Experimental Setup}

\textbf{Hardware.} NVIDIA Jetson Orin NX 16GB (100~TOPS: 60~TOPS GPU + $2 \times 20$~TOPS DLA), JetPack~6.2.1 (L4T~36.4.4), TensorRT~10.3.0, CUDA~12.6.

\textbf{Benchmarking protocol.} All latency measurements use CUDA event timing with 200\,ms warmup followed by a 30\,s timed run. Throughput is reported as queries per second (QPS) at the engine's static batch size of 16. Pipeline FPS measurements were repeated over 10 runs on two different Orin NX devices, and reported values are representative of these repeated measurements. All pipeline throughput measurements were collected at \textbf{25W power mode} on the Orin NX, as this profile represents a standard operational envelope for edge deployment. While MAXN SUPER mode would yield higher absolute throughput, testing at 25W provides a realistic benchmark under realistic power and thermal constraints.

\textbf{Metrics.} Classification accuracy, per-head ROC-AUC, per-head F1 score, confusion matrices, and GPU fallback layer count from TensorRT build logs. Both classification heads use a fixed decision threshold of $\tau{=}0.5$, the natural midpoint of the Sigmoid output range $[0,1]$, applied independently per head. This threshold is held constant across all backends (PyTorch, ONNX, TRT) to ensure a fair comparison.

\subsection{Accuracy Across Quantization Methods}

Table~\ref{tab:accuracy} compares accuracy and per-head ROC-AUC across quantization strategies.

\begin{table}[t]
\centering
\caption{Classification accuracy and per-head ROC-AUC across quantization methods. All DLA INT8 engines have zero GPU fallback layers.}
\label{tab:accuracy}
\begin{tabular}{@{}lccc@{}}
\toprule
\textbf{Method} & \textbf{Acc.} & \textbf{Hd-A AUC} & \textbf{Hd-B AUC} \\
\midrule
FP32 PyTorch                    & 94.5\% & 0.9863 & 0.9831 \\
FP32 ONNX                       & 94.5\% & 0.9863 & 0.9831 \\
QAT PyTorch (fake-quantized)     & 95.0\% & 0.9838 & 0.9777 \\
QAT ONNX (with Q/DQ)            & 95.0\% & 0.9839 & 0.9777 \\
\midrule
PTQ manual ranges (DLA INT8)    & 94.0\% & 0.9751 & 0.9728 \\
PTQ entropy (DLA INT8)          & 75.0\% & 0.8357 & 0.8590 \\
QAT percentile 99.99th (DLA INT8) & \textbf{95.0\%} & \textbf{0.9842} & \textbf{0.9779} \\
\bottomrule
\end{tabular}
\vspace{2pt}
\noindent\rule{\linewidth}{0.4pt}\par\vspace{1pt}
\noindent\parbox{\linewidth}{\footnotesize\raggedright\textit{FP32 ONNX and QAT ONNX match their PyTorch counterparts exactly, confirming lossless fused head export. QAT INT8 DLA matches or marginally exceeds the FP32 baseline, confirming that INT8 quantization introduces no meaningful accuracy degradation. Accuracy is the average across both heads at threshold $\tau{=}0.5$. Per-head macro-averaged F1 scores confirm that quantization preserves classification quality.}}
\end{table}

\textbf{Primary workflow: QAT DLA INT8.}
The recommended production path is FP32 training followed by QAT fine-tuning with 99.99th-percentile calibration, compiled to a DLA INT8 engine via explicit Q/DQ graph surgery. This aligns with TensorRT's own recommendation for explicit quantization~\cite{nvidia_pytorchquant} and delivers the strongest result, matching or marginally exceeding the FP32 baseline with zero GPU fallback (Table~\ref{tab:accuracy}).

\textbf{Experimental finding: PTQ manual ranges as a training-free approximation.}
During development we discovered that structurally derived manual dynamic ranges applied directly to the PTQ model without any QAT fine-tuning come within 1\,pp of the full QAT result on DLA INT8 (Table~\ref{tab:accuracy}). This emerged as a useful rapid-prototyping alternative because it bypasses the QAT training pipeline entirely, requires only minutes of engineering, and serves as a rapid deployment checkpoint to validate DLA operator compatibility and latency before committing QAT training time.

Table~\ref{tab:accuracy} also reveals an instructive metric discrepancy where PTQ manual achieves lower ROC-AUC than QAT DLA yet comparable accuracy. ROC-AUC is threshold-independent, measuring score \emph{ranking} quality across all thresholds, while accuracy is threshold-dependent at $\tau{=}0.5$. The gap is explained by a threshold calibration effect where INT8 quantization shifts absolute output values, displacing borderline samples across the $\tau{=}0.5$ decision boundary. QAT fine-tuning under quantization constraints corrects this by learning to keep output scores well separated around the threshold, recovering both well-calibrated decisions and better score distributions. In summary, PTQ manual ranges are suitable for experimentation and rapid prototyping, while QAT is preferred for production.

\subsection{Latency and Throughput}

Table~\ref{tab:latency} reports per-component latency for both DLA INT8 engines at batch size 16. DLA compute is near-identical across both engines, confirming that the choice of calibration method does not affect DLA execution time. The higher host-to-device latency on the QAT engine reflects the different input buffer layout produced by the Q/DQ graph surgery pipeline. Both engines achieve approximately 50\,QPS.

\begin{table}[t]
\centering
\caption{DLA INT8 engine latency breakdown (batch size = 16). PTQ and QAT engines have near-identical DLA compute. The higher H2D latency on QAT reflects the different input buffer layout produced by the Q/DQ graph surgery pipeline.}
\label{tab:latency}
\begin{tabular}{@{}lrr@{}}
\toprule
\textbf{Metric} & \textbf{PTQ manual} & \textbf{QAT percentile} \\
\midrule
Host-to-device (median) & 0.04\,ms & 0.44\,ms \\
DLA compute (median)    & 19.68\,ms & 19.71\,ms \\
Device-to-host (median) & $<$0.01\,ms & $<$0.01\,ms \\
\midrule
Total host latency (median) & 19.73\,ms & 20.16\,ms \\
Throughput              & 50.5\,QPS & 50.4\,QPS \\
\bottomrule
\end{tabular}
\vspace{2pt}
\noindent\rule{\linewidth}{0.4pt}\par\vspace{2pt}
\noindent\parbox{\linewidth}{\footnotesize\raggedright\textit{Since both engines have zero GPU fallback, this is entirely DLA execution time. Both engines are effectively identical in latency and throughput.}}
\end{table}

\subsection{Pipeline Throughput: The Value of DLA Offloading}

Table~\ref{tab:pipeline} demonstrates the core value proposition that classification on DLA is effectively free. Results are reported at two input resolutions to show the benefit across the operating range. In both cases the DLA classifier completes well within the YOLO frame interval. The parallel pipeline achieves near-identical FPS to YOLO alone, with the small overhead originating entirely from GPU-side person-crop preprocessing before DLA dispatch. The DLA execution itself runs fully concurrently and contributes zero latency to the primary detection path.

The sequential GPU baseline, running both YOLO and the classifier on the same GPU in series, incurs substantial throughput loss (Table~\ref{tab:pipeline}). The degradation is more severe at 720p because the classifier's fixed per-frame cost consumes a larger fraction of the shorter YOLO frame interval.

This reveals a counterintuitive result where the GPU FP16 classifier is faster than DLA in isolation, yet it produces worse pipeline throughput. The reason is that component latency is not the relevant metric for a shared-hardware pipeline. The classifier consumes GPU cycles that would otherwise be available to the detector, serializing two workloads on a single accelerator. DLA offloading eliminates this contention by executing classification on architecturally independent hardware, making it effectively free from the pipeline perspective.

\textbf{Scaling to two DLA classifiers.} To validate the multi-accelerator scaling claim, we deployed a second independent INT8 classifier on DLA core-1 alongside the existing classifier on DLA core-0, with YOLO detection on the GPU. The result (Table~\ref{tab:pipeline}) confirms that dual-DLA parallel throughput is identical to single-DLA parallel, as both DLA cores execute fully independently with zero mutual interference or GPU impact. In contrast, adding a second GPU FP16 classifier in the sequential configuration compounds the throughput loss further. Each additional DLA classifier adds zero pipeline overhead, while each additional GPU classifier adds serialized compute. This demonstrates that the methodology scales to the full heterogeneous SoC with GPU + DLA core~0 + DLA core~1, each processing independent workloads at full silicon utilization.

\begin{table}[t]
\centering
\caption{End-to-end pipeline throughput at two input resolutions (measured at 25W power mode on Orin NX).}
\label{tab:pipeline}
\begin{tabular}{@{}lr@{}}
\toprule
\textbf{Configuration} & \textbf{FPS} \\
\midrule
\multicolumn{2}{@{}l}{\small\textit{1080p input (high-resolution aerial imagery target)}} \\
\quad YOLO only (GPU)                              & 13.3          \\
\quad YOLO + 1 clf sequential (GPU)                & 10.5          \\
\quad YOLO + 2 clf sequential (GPU)                & 9.6           \\
\quad YOLO + 1 clf parallel (GPU+DLA0)             & \textbf{12.5} \\
\quad YOLO + 2 clf parallel (GPU+DLA0+DLA1)        & \textbf{12.5} \\
\addlinespace
\multicolumn{2}{@{}l}{\small\textit{720p input}} \\
\quad YOLO only (GPU)                              & 27.5          \\
\quad YOLO + 1 clf sequential (GPU)                & 19.0            \\
\quad YOLO + 2 clf sequential (GPU)                & 15.3          \\
\quad YOLO + 1 clf parallel (GPU+DLA0)             & \textbf{25.0} \\
\quad YOLO + 2 clf parallel (GPU+DLA0+DLA1)        & \textbf{25.0} \\
\bottomrule
\end{tabular}
\vspace{2pt}
\noindent\rule{\linewidth}{0.4pt}\par\vspace{1pt}
\noindent\parbox{\linewidth}{\footnotesize\raggedright\textit{Sequential throughput measured with GPU FP16 classifiers running after YOLO on the same GPU. FPS varies with detection count. Dual-DLA parallel uses DLA core~0 and DLA core~1 concurrently, each running an independent INT8 classifier on separate CUDA streams.}}
\end{table}

\subsection{Calibration Method Comparison}
\label{sec:calibration_comparison}

Table~\ref{tab:calibration} compares the three amax computation methods for QAT calibration on this model.

\begin{table}[h]
\centering
\caption{QAT calibration method comparison on DLA INT8 engines. Percentile achieves the highest threshold-aligned accuracy, while entropy and MSE achieve marginally higher AUC due to distribution-preserving scale choices.}
\label{tab:calibration}
\begin{tabular}{@{}lccc@{}}
\toprule
\textbf{Method} & \textbf{Acc.} & \textbf{Hd-A AUC} & \textbf{Hd-B AUC} \\
\midrule
Entropy (KL div.)    & 94.5\% & 0.9864          & \textbf{0.9832} \\
MSE                  & 94.5\% & \textbf{0.9866} & 0.9816          \\
Percentile (99.99th) & \textbf{95.0\%} & 0.9842 & 0.9779          \\
\bottomrule
\end{tabular}
\vspace{2pt}
\noindent\rule{\linewidth}{0.4pt}\par\vspace{1pt}
\noindent\parbox{\linewidth}{\footnotesize\raggedright\textit{AUC is threshold-independent (score \emph{ranking}), whereas accuracy is threshold-dependent at $\tau{=}0.5$. Percentile concentrates INT8 bin resolution on the dense $[0,6]$ ReLU6 range, keeping outputs calibrated around the decision boundary. Entropy and MSE optimize for distribution-preserving scales, improving global ranking at the cost of boundary alignment. For fixed-threshold deployment, percentile is the appropriate choice.}}
\end{table}

\begin{table*}[t]
\centering
\caption{Ablation study results. Each row removes one design decision from the final methodology.}
\label{tab:ablation}
\begin{tabular}{@{}cl>{\raggedright\arraybackslash}p{5.5cm}c@{}}
\toprule
\textbf{\#} & \textbf{Change} & \textbf{Effect} & \textbf{Ref} \\
\midrule
A1 & Entropy calib.\ instead of manual ranges (PTQ) & Accuracy $94.0\% \to 75\%$, INT8 saturation cascade & \hyperref[constr:c7]{C7} \\
\addlinespace
A2 & Export with Concat instead of fused head & DLA compiler crash: dimension assertion & \hyperref[constr:c3]{C3} \\
\addlinespace
A3 & Remove \texttt{skip\_connection\_quantizer} from skip connections & \textbf{FP16 on DLA:} 16 Add nodes in FP16, $126.98$\,ms ($6.4\times$ vs production). \textbf{GPU fallback:} 36 layers cascade to GPU (Add $+$ conv2 $+$ downsample), GPU occupied, pipeline parallelism lost & \hyperref[constr:c4]{C4} \\
\addlinespace
A4 & ReLU instead of ReLU6 & QAT accuracy drops 95\%$\to$94\% ($-$1pp), AUC $-$1.47pp/$-$1.37pp (Hd-A/Hd-B). \newline PTQ manual accuracy drops 94\%$\to$93\% ($-$1pp), AUC $-$1.0pp/$-$1.3pp. \newline PTQ entropy accuracy drops: Hd-A 73\%$\to$69\% ($-$4pp), Hd-B 77\%$\to$66\% ($-$11pp). \newline Latency unchanged (19.71\,ms) & \hyperref[constr:c7]{C7} \\
\addlinespace
A5 & Remove explicit \texttt{post\_backbone} and \texttt{post\_head} quantizers & All layers downstream of \texttt{layer4} assigned FP16 & \hyperref[constr:c5]{C5} \\
\bottomrule
\end{tabular}
\end{table*}

\subsection{Ablation Studies}
\label{sec:ablation}

Each ablation removes or changes one design decision to validate its necessity. Results for A1, A2, and A5 come from failures encountered during R\&D (\hyperref[tab:constraints]{Section~\ref*{sec:constraints}, Table~\ref*{tab:constraints}}). A3 and A4 results are from dedicated experiments described below.

\textbf{A3 - \texttt{skip\_connection\_quantizer} is a prerequisite for full INT8 DLA deployment.}
The ablation ONNX (77~Q/DQ pairs vs.\ 92 in production, where the 15-node gap reflects the 16 removed skip-connection quantizers offset by one for the fused head) produces two failure modes.

\emph{Failure mode 1- FP16 on DLA.} With \texttt{--fp16} permitted, TensorRT keeps all layers on DLA but assigns the 16 BasicBlock Add nodes (which lack calibrated INT8 scales) to FP16. DLA FP16 is substantially slower than INT8 where the median latency rises to $126.98$\,ms ($7.6$\,QPS), a $6.4\times$ regression versus production ($19.71$\,ms, $50.4$\,QPS). NVIDIA explicitly advises against FP16 on DLA for this reason.

\emph{Failure mode 2- GPU fallback.} Without \texttt{--fp16}, TensorRT falls back 36 layers to GPU because of the 16 Add nodes, the conv2 convolution feeding each add, and the downsample convolutions at strided blocks. Beyond the parallelism loss, each migrated layer incurs DLA-to-GPU memory transfer overhead that further increases latency. Regardless of the isolated latency of this configuration, any GPU fallback is a system-level failure for this application as the entire value proposition of DLA offloading is that the GPU remains free to execute the detection pipeline concurrently. Once any classifier layer runs on GPU, the two workloads compete for the same compute unit, the pipeline becomes sequential, and the throughput gain that motivated DLA deployment is negated.

Both outcomes confirm that \texttt{skip\_connection\_quantizer} on every skip connection is not an optional optimization but a structural requirement because without it there is no path to the zero-GPU-fallback INT8 DLA execution that enables true pipeline parallelism.

\textbf{A4 - ReLU6 improves accuracy and AUC at every operating point without latency cost.}
We initially trained with standard ReLU and switched to ReLU6 after observing consistent improvements across all three calibration methods (Table~\ref{tab:ablation}).

The advantage is architectural, not incidental. ReLU6's bounded output ($[0,6]$) produces denser, more uniform activation distributions that quantize with less information loss under INT8's 256 levels. This also makes manual range derivation principled because the $[-8,8]$ intermediate range is anchored by ReLU6's known bound plus residual-add headroom, whereas unbounded ReLU requires an unconstrained heuristic. The effect is most pronounced under TRT's implicit entropy calibration, where unbounded activations produce wider, sparser histograms that cause entropy to clip more aggressively, amplifying the INT8 saturation cascade. The practical recommendation for DLA INT8 deployment is to attempt manual dynamic ranges first regardless of activation function, as entropy calibration degrades accuracy in both bounded and unbounded cases, though the failure mechanism differs.

\subsection{Qualitative Results}

\begin{figure*}[t]
    \centering
    \begin{subfigure}[b]{0.24\textwidth}
        \includegraphics[width=\textwidth]{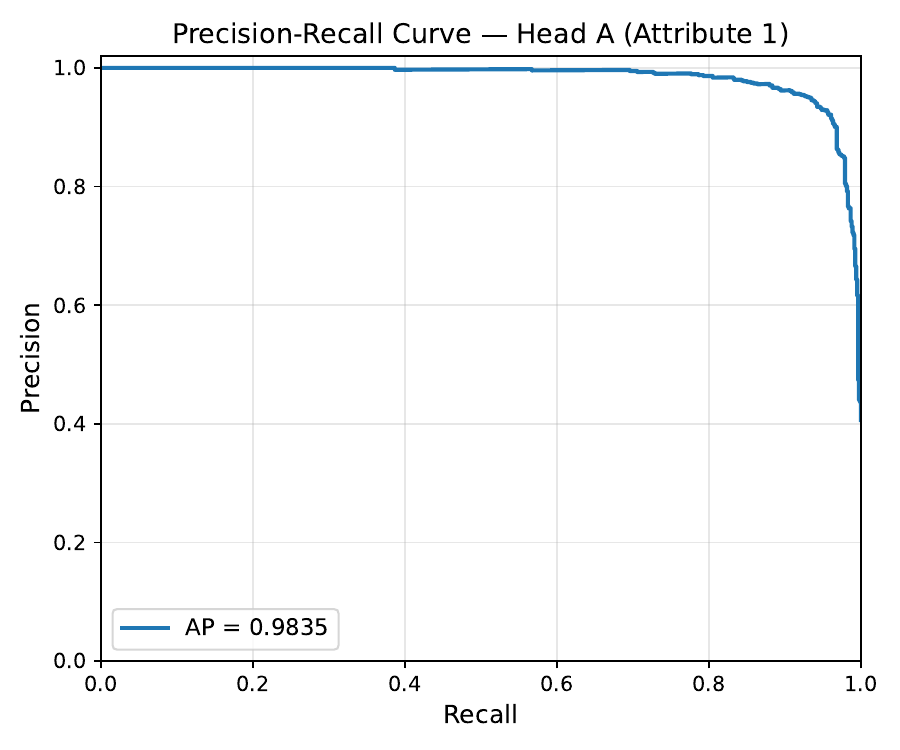}
        \caption{FP32- Head A}
    \end{subfigure}
    \hfill
    \begin{subfigure}[b]{0.24\textwidth}
        \includegraphics[width=\textwidth]{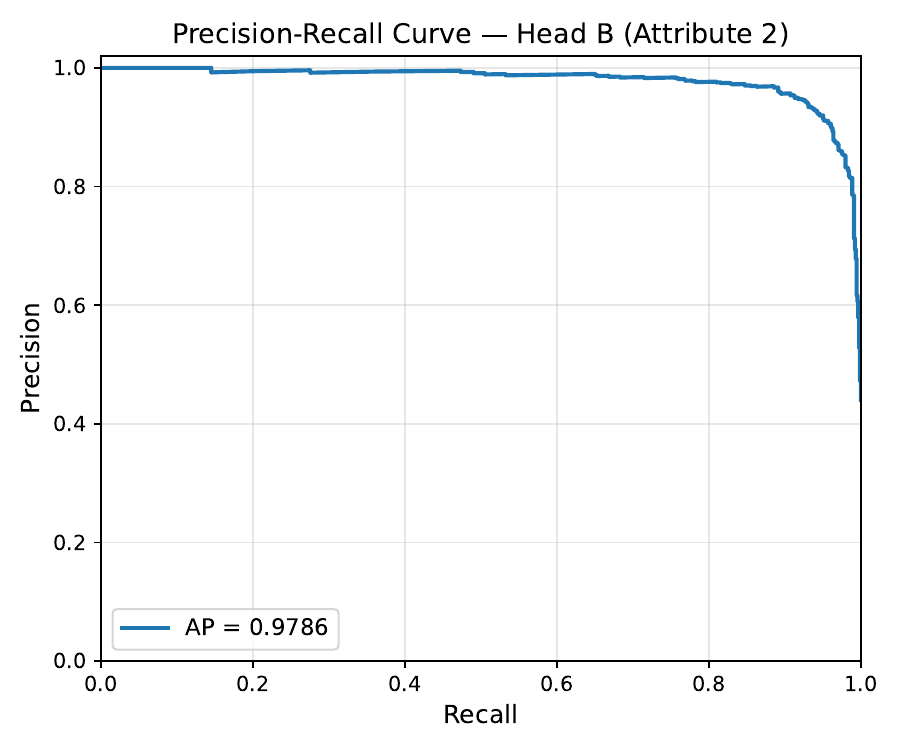}
        \caption{FP32- Head B}
    \end{subfigure}
    \hfill
    \begin{subfigure}[b]{0.24\textwidth}
        \includegraphics[width=\textwidth]{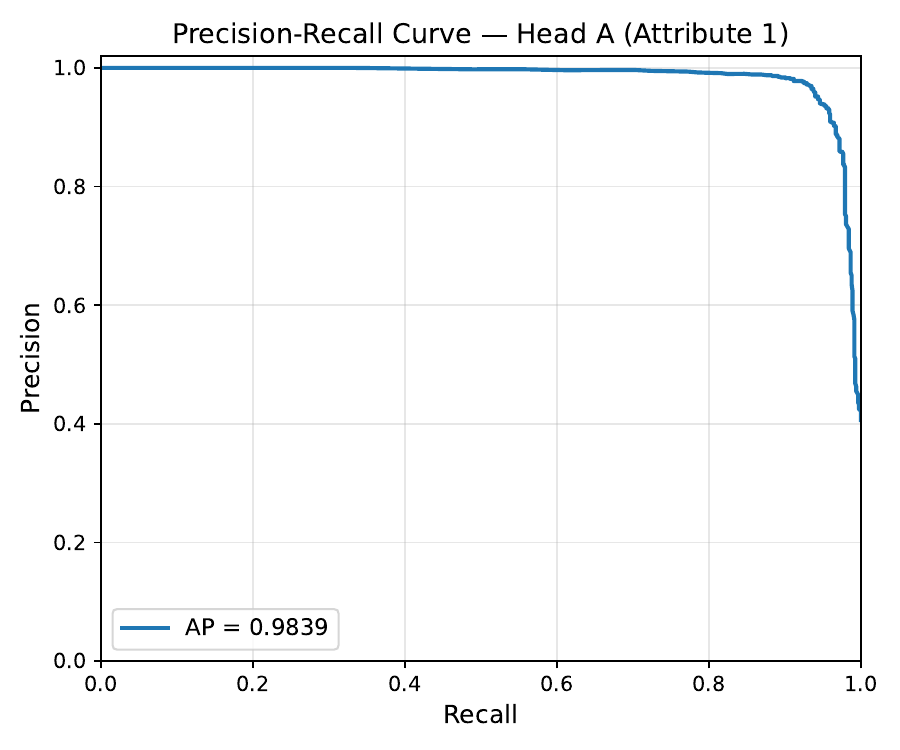}
        \caption{DLA INT8- Head A}
    \end{subfigure}
    \hfill
    \begin{subfigure}[b]{0.24\textwidth}
        \includegraphics[width=\textwidth]{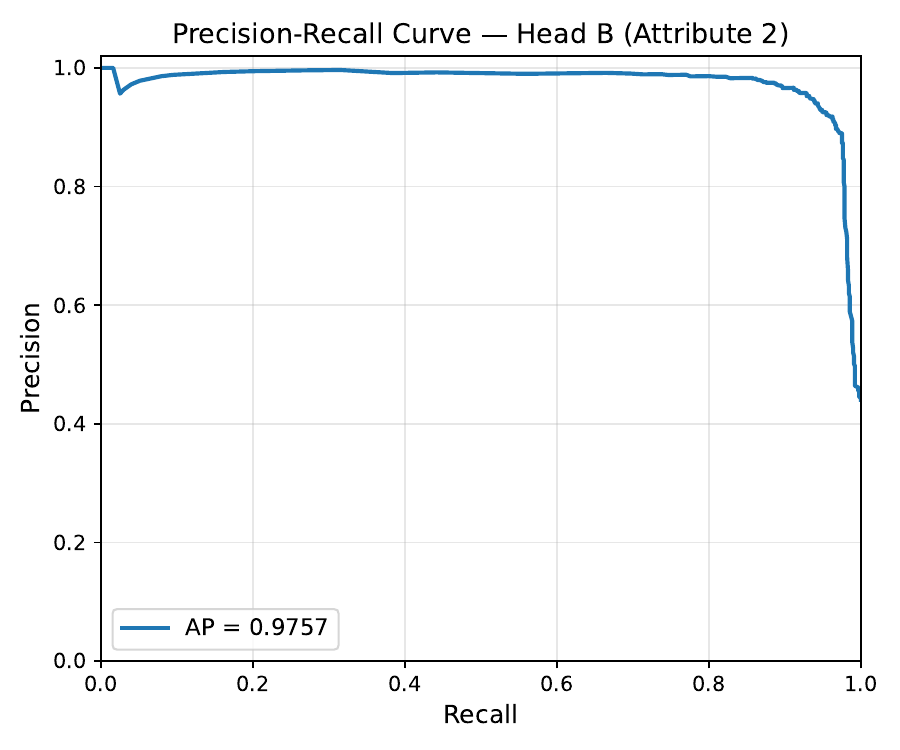}
        \caption{DLA INT8- Head B}
    \end{subfigure}
    \caption{Precision-Recall curves for FP32 and QAT DLA INT8. FP32 average precision: Head~A AP~=~0.9835, Head~B AP~=~0.9786. DLA INT8 average precision: Head~A AP~=~0.9839, Head~B AP~=~0.9757. Curves overlap closely, confirming quantization does not degrade precision-recall trade-offs.}
    \label{fig:pr}
\end{figure*}

Figures~\ref{fig:confusion}--\ref{fig:pr} present confusion matrices, ROC curves, and precision--recall curves for both FP32 and QAT DLA INT8 configurations.

The 4-class confusion matrices (Fig.~\ref{fig:confusion}a--b) reconstruct joint predictions from the two independent binary heads, where each cell represents one of the four joint attribute combinations. The dual-head architecture produces 4-class discrimination without a multi-class softmax where each head is independently calibrated and the joint prediction is their product. Classification patterns are near-identical between FP32 and DLA INT8. The per-head binary confusion matrices (Fig.~\ref{fig:confusion_perhead}) confirm this at the individual head level as both heads closely maintain their true-positive and true-negative counts after quantization, with no systematic shift toward either class.

ROC curves (Fig.~\ref{fig:roc}) show that INT8 quantization preserves ranking quality with negligible AUC differences across both heads (Table~\ref{tab:accuracy}). Precision--recall curves (Fig.~\ref{fig:pr}) similarly overlap across the full recall range, confirming that the operating-point trade-off is unchanged after quantization.

\noindent\begin{minipage}{\columnwidth}
    \centering
    \begin{minipage}[b]{0.49\columnwidth}
        \centering
        \includegraphics[width=\textwidth]{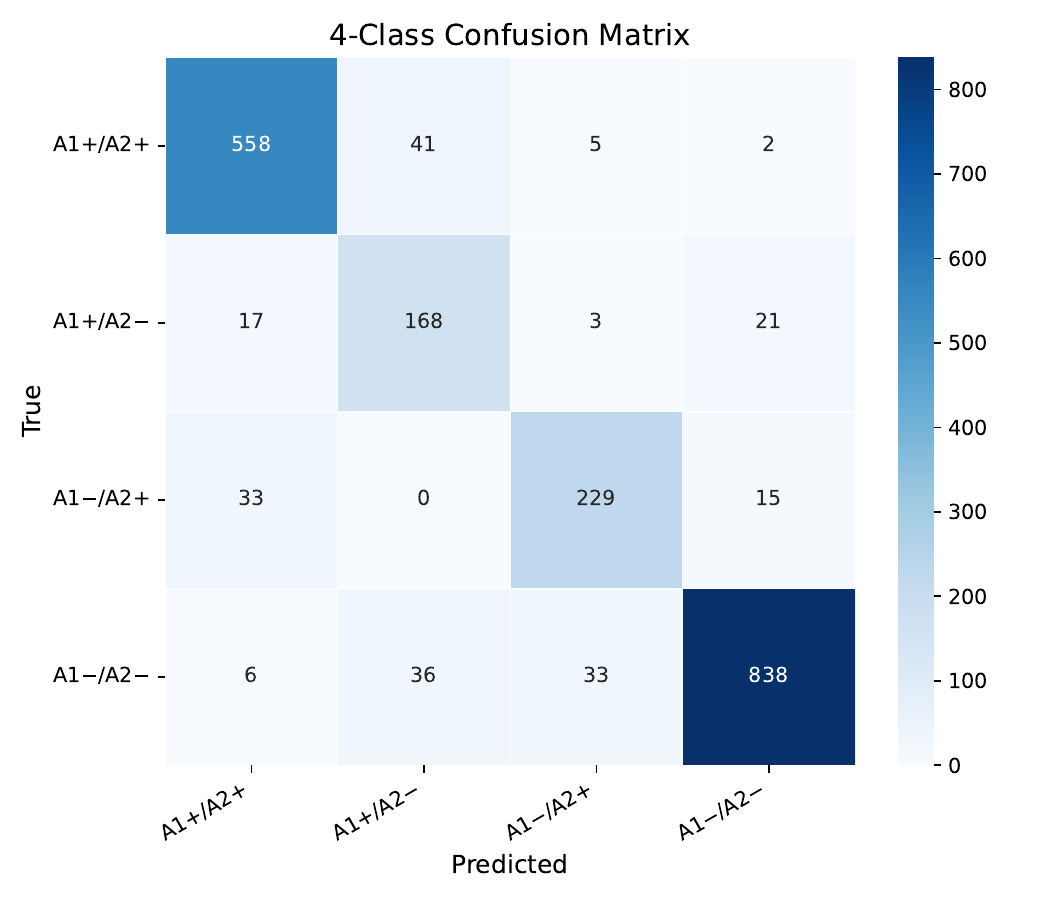}\\
        {\footnotesize (a) FP32 PyTorch}
    \end{minipage}
    \hfill
    \begin{minipage}[b]{0.49\columnwidth}
        \centering
        \includegraphics[width=\textwidth]{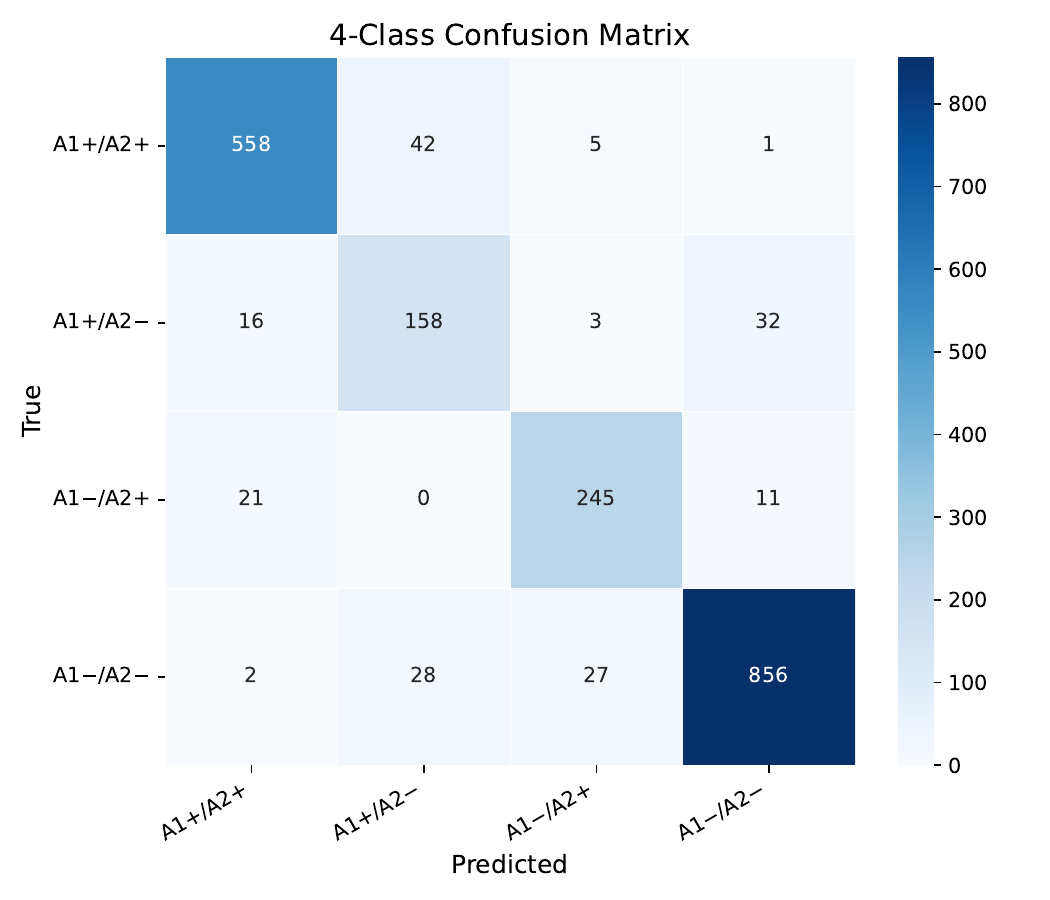}\\
        {\footnotesize (b) QAT DLA INT8}
    \end{minipage}
    \vspace{4pt}
    \captionof{figure}{4-class confusion matrices (FP32 vs.\ QAT DLA INT8).}
    \label{fig:confusion}
\end{minipage}

\vspace{6pt}

\noindent\begin{minipage}{\columnwidth}
    \centering
    \begin{minipage}[b]{0.48\columnwidth}
        \centering
        \includegraphics[width=\textwidth]{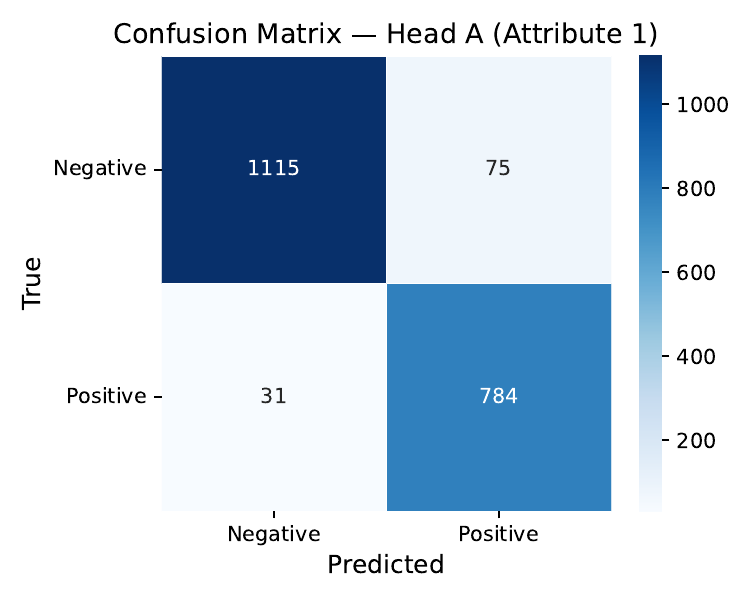}\\
        {\footnotesize (a) FP32 — Head A}
    \end{minipage}
    \hfill
    \begin{minipage}[b]{0.48\columnwidth}
        \centering
        \includegraphics[width=\textwidth]{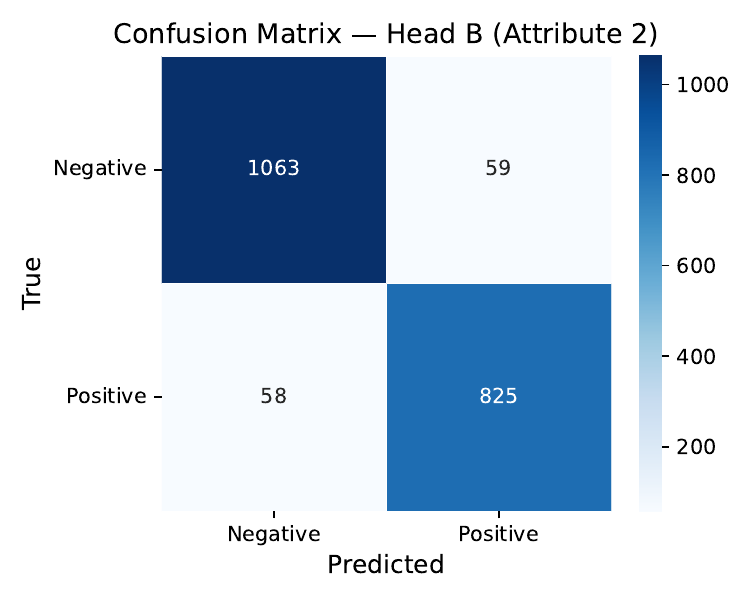}\\
        {\footnotesize (b) FP32 — Head B}
    \end{minipage}
    \vspace{4pt}
    \begin{minipage}[b]{0.48\columnwidth}
        \centering
        \includegraphics[width=\textwidth]{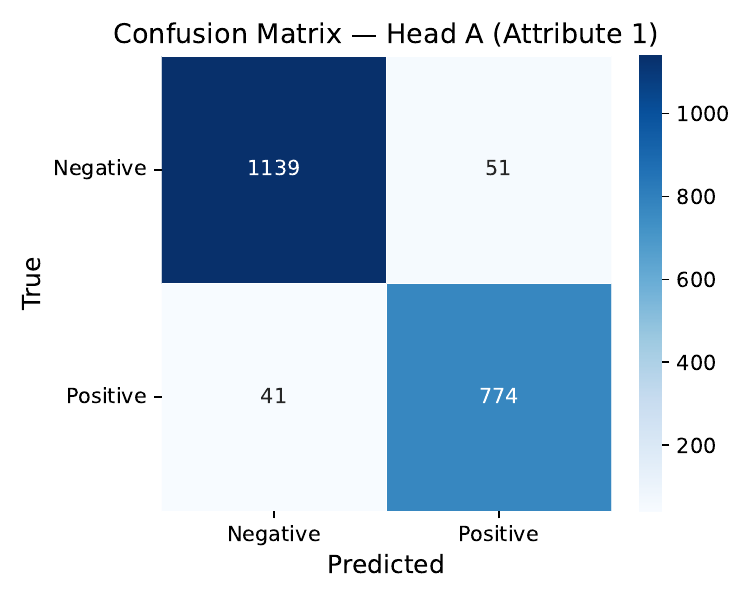}\\
        {\footnotesize (c) DLA INT8 — Head A}
    \end{minipage}
    \hfill
    \begin{minipage}[b]{0.48\columnwidth}
        \centering
        \includegraphics[width=\textwidth]{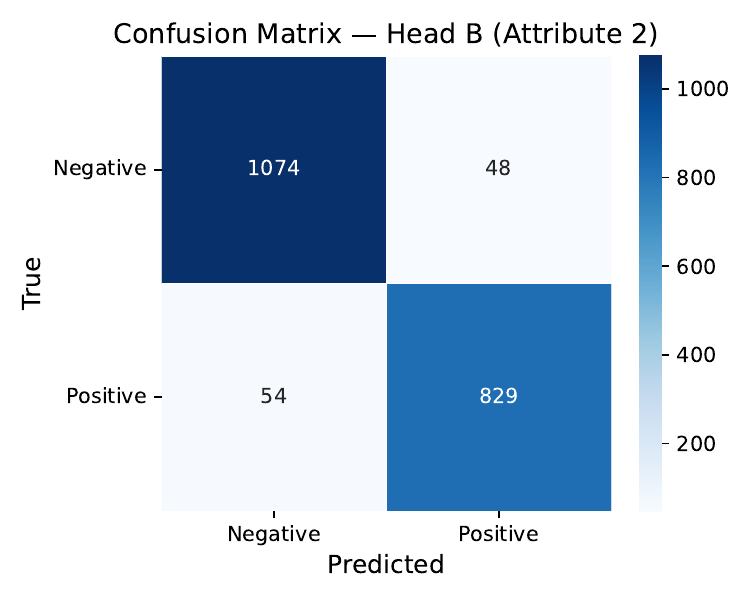}\\
        {\footnotesize (d) DLA INT8 — Head B}
    \end{minipage}
    \vspace{4pt}
    \captionof{figure}{Per-head binary confusion matrices (Negative/Positive) for FP32 and QAT DLA INT8. Each head is evaluated independently at threshold $\tau{=}0.5$.}
    \label{fig:confusion_perhead}
\end{minipage}

\vspace{6pt}

\noindent\begin{minipage}{\columnwidth}
    \centering
    \begin{minipage}[b]{0.48\columnwidth}
        \centering
        \includegraphics[width=\textwidth]{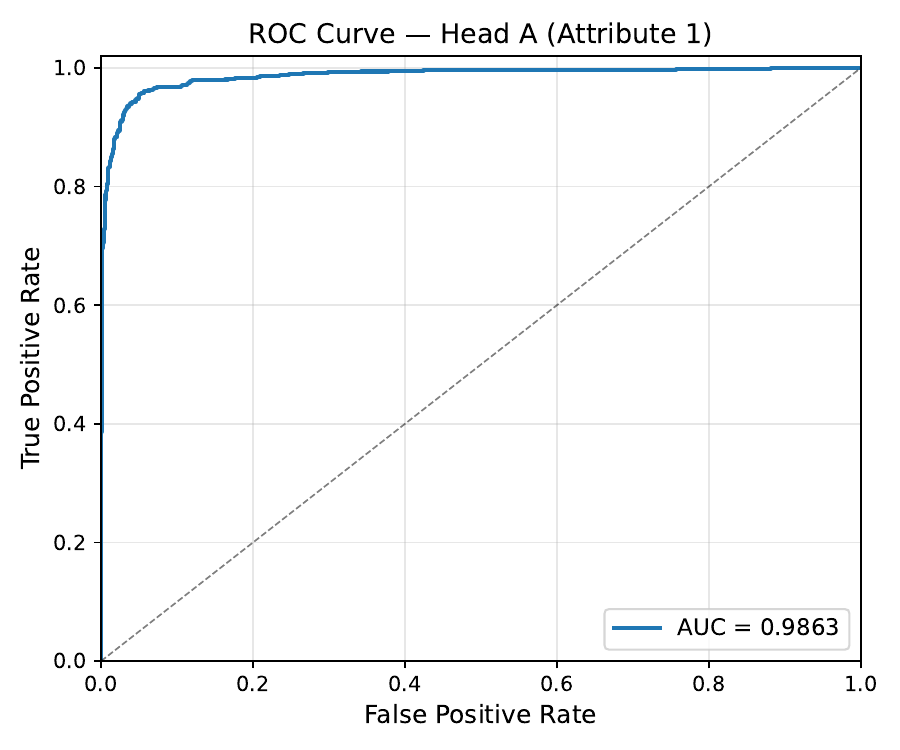}\\
        {\footnotesize (a) FP32 Head A}
    \end{minipage}
    \hfill
    \begin{minipage}[b]{0.48\columnwidth}
        \centering
        \includegraphics[width=\textwidth]{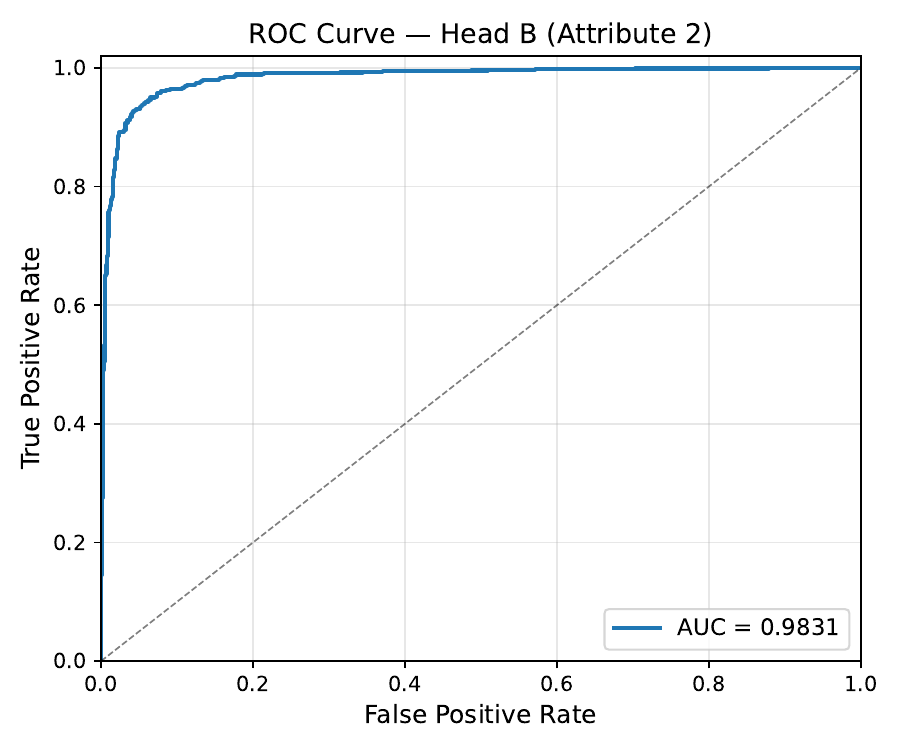}\\
        {\footnotesize (b) FP32 Head B}
    \end{minipage}
    \vspace{4pt}
    \begin{minipage}[b]{0.48\columnwidth}
        \centering
        \includegraphics[width=\textwidth]{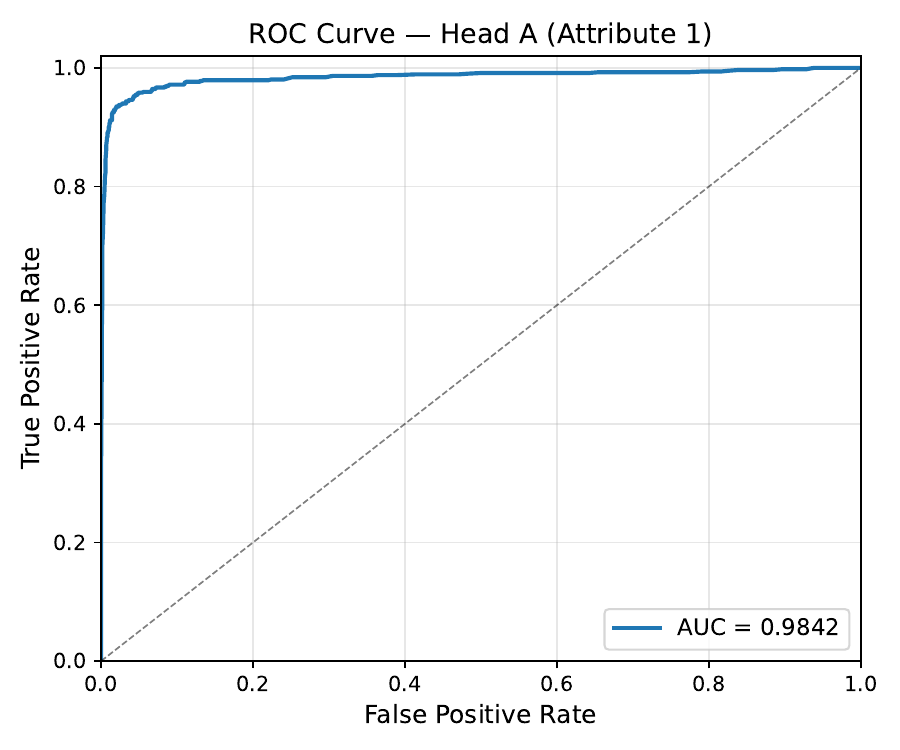}\\
        {\footnotesize (c) DLA INT8 Head A}
    \end{minipage}
    \hfill
    \begin{minipage}[b]{0.48\columnwidth}
        \centering
        \includegraphics[width=\textwidth]{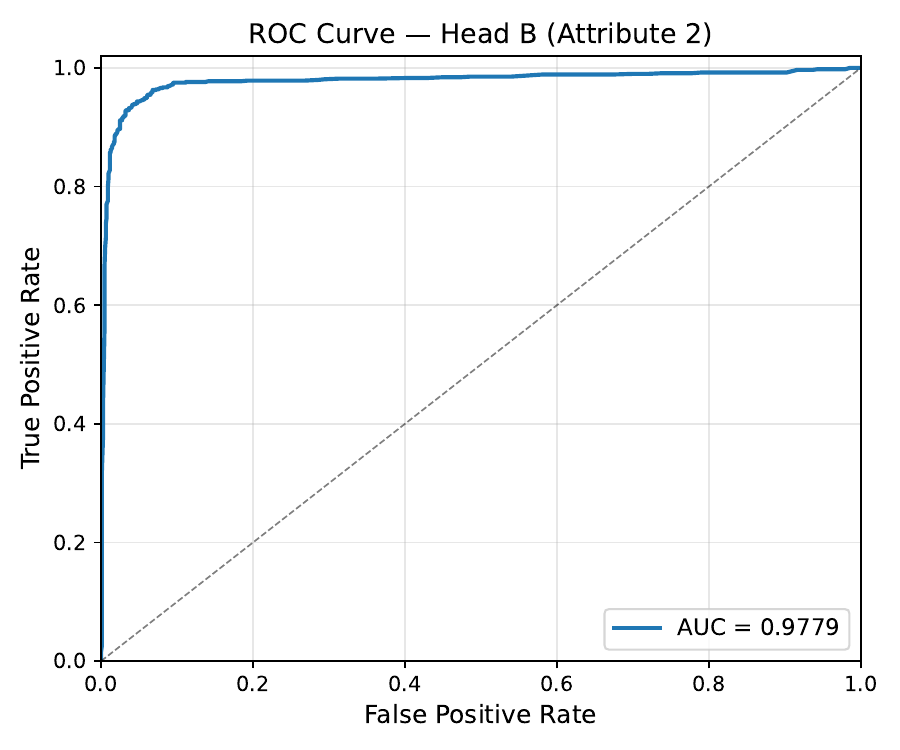}\\
        {\footnotesize (d) DLA INT8 Head B}
    \end{minipage}
    \vspace{4pt}
    \captionof{figure}{ROC curves for FP32 and QAT DLA INT8. FP32 baseline: Head~A AUC~=~0.9863, Head~B AUC~=~0.9831. DLA INT8: Head~A AUC~=~0.9842, Head~B AUC~=~0.9779. Curve shapes are near-identical, confirming INT8 quantization preserves ranking quality.}
    \label{fig:roc}
\end{minipage}

\section{Discussion}
\label{sec:discussion}

\subsection{Generalization: When Does This Methodology Apply?}

The five-step methodology is not specific to ResNet-34 or person attribute classification. It applies whenever

\begin{enumerate}[itemsep=2pt]
    \item The target SoC has DLA cores (all NVIDIA Jetson variants from Xavier onwards).
    \item A classification model must run concurrently with a GPU workload, or DLA compute is available and underutilized.
    \item The classification backbone can be adapted with the operator replacements in Step~1. Residual architectures (ResNet-18/34/50) are directly applicable. However, lightweight architectures with depthwise separable convolutions (MobileNetV2/V3, EfficientNet-Lite) generalize in principle but require per-operator DLA verification before deployment.
\end{enumerate}

Example applications beyond ATR include vehicle type and make classification in traffic systems, PPE (helmet, vest) detection in industrial safety, face attribute analysis, medical image triage (normal/abnormal screening), agricultural crop classification from drone imagery, and quality inspection in manufacturing. In each case, the detection model runs on GPU while the classifier runs on DLA. The specific classification task changes as per requirements, but the deployment methodology is identical.

Although demonstrated on NVIDIA DLA, the asynchronous pipeline architecture generalizes to any heterogeneous edge SoC that pairs a primary compute unit with a dedicated neural accelerator. Examples include Qualcomm platforms (Adreno GPU + Hexagon NPU), Ambarella CVflow-based SoCs, and Texas Instruments TDA4 (C7x DSP + MMA accelerator). In each case, the detection model runs on the primary compute unit while classifiers offload to the dedicated accelerator, achieving the same frame-level parallelism without contention on the primary compute path.

\subsection{Why DLA Offloading Achieves Near-Zero Overhead and Why GPU Multi-Stream Cannot}

A natural question is whether the same $N{-}1$ parallel architecture could be achieved by running the classifier on the GPU alongside the detector using CUDA multi-stream execution. The answer is yes in principle, but the result is not zero overhead.

CUDA multi-stream allows two kernels to be enqueued concurrently, but both compete for the same physical resources, namely streaming multiprocessors (SMs), L2 cache, and DRAM bandwidth. While we validate with YOLOv11m, the argument strengthens as GPU-side models grow larger. Transformer-based detectors such as RT-DETR, RF-DETR, RTMDet, etc., and future architectures with higher parameter counts and attention mechanisms consume even more GPU resources, leaving progressively fewer SMs available for a concurrent classifier. The classifier's kernels cannot be guaranteed to execute in parallel. The SM and memory bandwidth contention means they are likely queued and delayed behind the detector~\cite{cuda_programming_guide,majeed2026scheduling,tayal2025multiinstance}. Any GPU cycles consumed by the classifier are cycles unavailable to the detector, and the overhead scales with both classifier and detector workload. The measured sequential GPU baseline (Table~\ref{tab:pipeline}) quantifies this serial cost for YOLOv11m, and multi-stream execution is unlikely to improve significantly upon it, as SM saturation leaves little room for concurrent execution. For heavier GPU models the contention would be even more severe, making DLA offloading increasingly valuable.

DLA is architecturally separate hardware with its own compute engines, local SRAM, and power domain, connected to the SoC fabric independently of the GPU. When DLA executes the classifier, the GPU's SMs, caches, and memory bandwidth are completely unaffected. The detector throughput is identical whether the classifier is running or not, regardless of detector model size. This is not merely pipelining across frames, but true hardware-level parallelism that imposes near-zero overhead on the primary detection path. The methodology is detector-agnostic and applies to any GPU workload.

Additionally, DLA's INT8 specialized compute delivers 3--5$\times$ higher performance per watt compared to GPU execution for the same workload~\cite{nvidia_dla_perf}.

\subsection{Full SoC Utilization}

The Orin NX provides 100~TOPS, with 60~TOPS on the GPU and $2 \times 20$~TOPS across two DLA cores~\cite{nvidia_orin_datasheet}. In GPU-only deployments, 40\% of purchased compute sits idle. For cost-sensitive edge platforms such as drones, mobile systems, and distributed sensor networks, this is not merely an efficiency concern but a direct cost multiplier, as idle silicon represents paid-for compute that delivers no return.

With DLA-deployable models, the full heterogeneous architecture is utilized.

\begin{itemize}[itemsep=2pt]
    \item \textbf{Parallel pipeline.} The GPU executes detection while the DLA executes classification concurrently. The measured overhead originates entirely from GPU-side crop preprocessing, not from DLA execution (Table~\ref{tab:pipeline}).
    \item \textbf{Multi-accelerator scaling.} GPU + DLA core~0 + DLA core~1 each run independent models concurrently. The second DLA classifier adds zero overhead, whereas a second GPU classifier compounds sequential throughput loss (Table~\ref{tab:pipeline}).
    \item \textbf{Across the Jetson family.} Orin Nano (GPU + 1~DLA), Orin NX (GPU + 2~DLA), AGX Orin (GPU + 2~DLA at higher TOPS). The methodology scales across all variants.
\end{itemize}

This reframes DLA from a demo feature to a production compute resource representing 40\% of the SoC's total capability.

\subsection{Extensibility: Adding Classification Heads}

The same method used to construct the dual-head architecture generalizes to any number of heads. Additional attribute classifiers (e.g., pose estimation, equipment type, vehicle make) can be appended, trained and then fused into the single Conv2d during export. Each additional head adds minimal DLA cost, as a $1\times1$ convolution with one output channel is a single DLA-native operation.

\subsection{Limitations}

\begin{enumerate}[itemsep=2pt]
    \item \textbf{Static batch size:} DLA engines require a fixed batch size. When the number of detections per frame is fewer than the batch size, the input must be padded with zero tensors, wasting some DLA compute cycles.

    \item \textbf{Frame $N{-}1$ latency:} The asynchronous pipeline means classification results always lag detection by exactly one frame. This is acceptable for ATR and surveillance applications where targets move predictably between consecutive frames.
    \item \textbf{Multi-head gradient entanglement:} When certain attribute combinations have insufficient training data, gradient detachment (\texttt{detach\_head\_b}) prevents the data-scarce head from degrading the backbone. However, this limits the data-scarce head's ability to benefit from backbone fine-tuning.

    \item \textbf{No comparison with alternative frameworks:} We do not compare against TensorFlow Lite, ONNX Runtime, or NVIDIA DeepStream. TensorRT delivers the highest throughput and accuracy on Jetson hardware, making it the natural choice for DLA deployment. DeepStream provides DLA integration for NVIDIA's sample models but does not expose the custom model adaptation pipeline (architecture surgery, explicit Q/DQ insertion, graph surgery for calibration cache generation) that this work addresses. The methodology presented here produces DLA-ready engines that can be integrated into DeepStream or any other serving framework.
\end{enumerate}

\section{Future Work}

\label{sec:future_work}

\subsection{2:4 Structured Sparsity on DLA}

DLA on Orin supports 2:4 structured sparsity for INT8 convolutions with more than 64 output channels, offering approximately $2\times$ throughput at near-identical accuracy. Because all layers in our proposed ResNet-34 backbone have $\geq 64$ channels, the model fully qualifies for this optimization. The process involves applying 2:4 sparse masks to trained weights, fine-tuning for 5--10 epochs to recover accuracy, and rebuilding the engine. This could approximately halve the current batch-16 inference latency (Table~\ref{tab:latency}). Furthermore, the computational headroom unlocked by this sparsity can be reinvested into a higher-capacity backbone such as ResNet-50, to improve classification accuracy. Because a sparse ResNet-50 leverages the same hardware acceleration, it can deliver better accuracy while maintaining an inference latency comparable to the unoptimized ResNet-34 baseline.

\subsection{Backbone Scaling}

The methodology applies directly to other ResNet variants:

\begin{itemize}[itemsep=2pt]
    \item \textbf{ResNet-50} (25.6M parameters): Uses bottleneck blocks with $1\times1$ and $3\times3$ convolutions, which are DLA-safe. It features a higher accuracy ceiling but offers lower raw throughput.

    \item \textbf{ResNet-18} (11.7M parameters): Lighter and faster, suitable when the latency budget is tighter than accuracy requirements and runs alongside a lighter GPU detector.

    \item \textbf{DenseNet-121~\cite{huang2017densenet}} (8.0M parameters): Dense connectivity promotes feature reuse and gradient flow, achieving competitive accuracy with fewer parameters than ResNet-34. All operations (BN-ReLU-Conv $1\times1$/$3\times3$, average pooling, concatenation) are DLA-compatible, making it a strong candidate for higher accuracy within the same latency envelope.

    \item \textbf{MobileNetV2/V3~\cite{sandler2018mobilenetv2, howard2019mobilenetv3}, EfficientNet-Lite~\cite{tan2019efficientnet}}: Lighter architectures with depthwise separable convolutions. While the DLA natively supports depthwise \texttt{Conv2d} layers, these models still require per-operator verification.
\end{itemize}

\subsection{Standalone DLA Mode}

Current deployment uses Standard DLA mode with GPU fallback enabled (though zero layers fall back) for IO reformat before feeding the GPU processed image frame tensors to the DLA core. This reformat is a negligible overhead as the GPU handles the tensor format conversion natively rather than performing it on the CPU during inference. Standalone mode (\texttt{kDLA\_STANDALONE} with \texttt{kDIRECT\_IO}) eliminates the GPU context entirely for faster cold-start and dedicated DLA execution. This requires reformat-free I/O (\texttt{int8:dla\_hwc4} input, \texttt{int8:chw32} output), meaning the inference pipeline must deliver pre-formatted INT8 tensors.

\subsection{Channel Pruning with Sparsity}

Pruning 30--50\% of channels before applying 2:4 sparsity could compound throughput gains. The explicit quantization pipeline handles pruned architectures identically because \texttt{QuantConv2d} adapts to any channel width. The risk is that aggressive pruning may reduce classification accuracy below operational thresholds.

\section{Conclusion}
\label{sec:conclusion}

We presented a systematic five-step methodology for deploying custom classification models on NVIDIA DLA with zero GPU fallback, enabling parallel multi-model inference on a single edge SoC. The methodology covers architecture adaptation (DLA-safe operator replacement and multi-head fusion), explicit INT8 quantization with per-layer Q/DQ coverage, manual dynamic ranges, QAT with configurable calibration methods, and DLA-safe ONNX graph surgery to produce a calibration cache for the TensorRT DLA engine build. We documented nine engineering constraints encountered during development, each representing a previously undocumented failure mode accompanied by root-cause analysis and a generalizable solution. Notably, manual dynamic ranges enable rapid end-to-end validation of the DLA INT8 pipeline without requiring QAT, allowing practitioners to verify correct deployment before investing in quantization-aware fine-tuning.

Validation on a dual-head person attribute classifier deployed alongside a GPU-based detector (YOLOv11m in our validation) on an NVIDIA Jetson Orin NX demonstrates that the frame $N{-}1$ parallel architecture makes classification effectively free (Table~\ref{tab:pipeline}). The measured pipeline overhead originates from crop preprocessing, not DLA execution.

Beyond parallel pipelines, this work enables full utilization of the Jetson SoC's heterogeneous compute. Deploying classifiers on both DLA cores alongside GPU detection achieves identical throughput to a single-DLA configuration, while adding a second GPU classifier compounds sequential throughput loss (Table~\ref{tab:pipeline}).

The methodology supports arbitrary classification heads and is both backbone-agnostic and detector-agnostic, applying to any detection-classification pipeline across defense, surveillance, autonomous vehicles, industrial inspection, and beyond. The benefit of DLA offloading scales with GPU model complexity, as heavier detectors leave even less GPU capacity for concurrent classifiers, making DLA offloading increasingly valuable. This work fills a critical gap between NVIDIA's hardware capabilities and the practical engineering knowledge required to exploit them.

\bibliographystyle{IEEEtran}
\bibliography{references}

\begin{thebibliography}{10}
\providecommand{\url}[1]{#1}
\csname url@samestyle\endcsname
\providecommand{\newblock}{\relax}
\providecommand{\bibinfo}[2]{#2}
\providecommand{\BIBentrySTDinterwordspacing}{\spaceskip=0pt\relax}
\providecommand{\BIBentryALTinterwordstretchfactor}{4}
\providecommand{\BIBentryALTinterwordspacing}{\spaceskip=\fontdimen2\font plus
\BIBentryALTinterwordstretchfactor\fontdimen3\font minus
  \fontdimen4\font\relax}
\providecommand{\BIBforeignlanguage}[2]{{%
\expandafter\ifx\csname l@#1\endcsname\relax
\typeout{** WARNING: IEEEtran.bst: No hyphenation pattern has been}%
\typeout{** loaded for the language `#1'. Using the pattern for}%
\typeout{** the default language instead.}%
\else
\language=\csname l@#1\endcsname
\fi
#2}}
\providecommand{\BIBdecl}{\relax}
\BIBdecl

\bibitem{nvidia_orin_datasheet}
{NVIDIA Corporation}, ``Jetson orin nx series data sheet,''
  \url{https://developer.nvidia.com/embedded/jetson-orin-nx}, 2023, 100 TOPS:
  60 GPU + 2$\times$20 DLA.

\bibitem{nvidia_dla_docs}
------, ``Nvidia deep learning accelerator,''
  \url{https://docs.nvidia.com/deeplearning/tensorrt/developer-guide/index.html},
  2024, tensorRT Developer Guide, DLA chapter.

\bibitem{tayal2025multiinstance}
M.~Tayal and Y.~Simmhan, ``Evaluating multi-instance dnn inferencing on
  multiple accelerators of an edge device,'' in \emph{2024 IEEE 31st
  International Conference on High Performance Computing, Data and Analytics
  Workshop (HiPCW)}.\hskip 1em plus 0.5em minus 0.4em\relax IEEE, 2024, pp.
  181--182.

\bibitem{majeed2026scheduling}
A.~A. Majeed, M.~Meribout, and S.~M. Sali, ``Scheduling techniques of ai models
  on modern heterogeneous edge gpu—a critical review,'' \emph{IEEE
  Transactions on Industrial Informatics}, vol.~22, no.~4, pp. 2641--2652,
  2026.

\bibitem{cpcnn2023}
D.~Chun, J.~Choi, H.-J. Lee, and H.~Kim, ``Cp-cnn: Computational
  parallelization of cnn-based object detectors in heterogeneous embedded
  systems for autonomous driving,'' \emph{IEEE Access}, vol.~11, pp.
  52\,812--52\,823, 2023.

\bibitem{nvidia_jetson_dla_tutorial}
{NVIDIA-AI-IOT}, ``Getting started with the deep learning accelerator on nvidia
  jetson orin,'' \url{https://github.com/NVIDIA-AI-IOT/jetson_dla_tutorial},
  2022, official end-to-end tutorial for DLA-compatible model adaptation.

\bibitem{nvidia_trt_work_with_dla}
{NVIDIA Corporation}, ``Working with dla in tensorrt,''
  \url{https://docs.nvidia.com/deeplearning/tensorrt/latest/inference-library/work-with-dla.html},
  2025, dLA support, restrictions, GPU fallback behavior, and deployment APIs.

\bibitem{gholami2021survey}
A.~Gholami, S.~Kim, D.~Zhen, Z.~Yao, M.~Mahoney, and K.~Keutzer, ``A survey of
  quantization methods for efficient neural network inference,'' in
  \emph{Low-Power Computer Vision: Improving the Efficiency of Artificial
  Intelligence}.\hskip 1em plus 0.5em minus 0.4em\relax Taylor \& Francis,
  2022, pp. 291--326.

\bibitem{jacob2018quantization}
B.~Jacob, S.~Kligys, B.~Chen, M.~Zhu, M.~Tang, A.~Howard, H.~Adam, and
  D.~Kalenichenko, ``Quantization and training of neural networks for efficient
  integer-arithmetic-only inference,'' in \emph{Proceedings of the IEEE
  Conference on Computer Vision and Pattern Recognition (CVPR)}, 2018, pp.
  2704--2713.

\bibitem{liang2021pruning}
T.~Liang, J.~Glossner, L.~Wang, S.~Shi, and X.~Zhang, ``Pruning and
  quantization for deep neural network acceleration: A survey,''
  \emph{Neurocomputing}, vol. 461, pp. 370--403, 2021.

\bibitem{nagel2021whitepaper}
M.~Nagel, R.~A. Amjad, M.~van Baalen, C.~Louizos, and T.~Blankevoort, ``A white
  paper on neural network quantization,'' \emph{arXiv preprint
  arXiv:2106.08295}, 2021.

\bibitem{esser2020lsq}
\BIBentryALTinterwordspacing
S.~K. Esser, J.~L. McKinstry, D.~Bablani, R.~Appuswamy, and D.~S. Modha,
  ``Learned step size quantization,'' in \emph{Proceedings of the International
  Conference on Learning Representations (ICLR)}, 2020. [Online]. Available:
  \url{https://openreview.net/forum?id=rkgO66VKDS}
\BIBentrySTDinterwordspacing

\bibitem{banner2019posttraining}
\BIBentryALTinterwordspacing
R.~Banner, Y.~Nahshan, and D.~Soudry, ``Post training 4-bit quantization of
  convolutional networks for rapid-deployment,'' \emph{Advances in Neural
  Information Processing Systems (NeurIPS)}, vol.~32, 2019. [Online].
  Available:
  \url{https://proceedings.neurips.cc/paper/2019/hash/c0a62e133894cdce435bcb4a5df1db2d-Abstract.html}
\BIBentrySTDinterwordspacing

\bibitem{chae2025pruneqd}
W.~{Chae} and K.~{Seo}, ``Pipeline of pruning, knowledge distillation, and
  quantization for model compression,'' \emph{Journal of Electrical Engineering
  \& Technology}, vol.~21, 12 2025.

\bibitem{huang2024coreset}
\BIBentryALTinterwordspacing
X.~Huang, Z.~Liu, S.-Y. Liu, and K.-T. Cheng, ``Robust and efficient
  quantization-aware training via coreset selection,'' \emph{Transactions on
  Machine Learning Research}, 2024. [Online]. Available:
  \url{https://openreview.net/forum?id=4c2pZzG94y}
\BIBentrySTDinterwordspacing

\bibitem{safdar2023yoloatr}
M.~Safdar \emph{et~al.}, ``Yoloatr: Thermal ir target recognition,'' in
  \emph{25th Irish Machine Vision and Image Processing Conference (IMVIP)},
  Galway, Ireland, 2023.

\bibitem{wang2023weapon}
G.~Wang, H.~Ding, M.~Duan, Y.~Pu, Z.~Yang, and H.~Li, ``Fighting against
  terrorism: A real-time cctv autonomous weapons detection based on improved
  {YOLO} v4,'' \emph{Digital Signal Processing}, vol. 132, p. 103790, 2023.

\bibitem{wang2022par}
\BIBentryALTinterwordspacing
X.~Wang, S.~Zheng, R.~Yang, A.~Zheng, Z.~Chen, J.~Tang, and B.~Luo,
  ``Pedestrian attribute recognition: A survey,'' \emph{Pattern Recognition},
  vol. 121, p. 108220, 2022. [Online]. Available:
  \url{https://www.sciencedirect.com/science/article/pii/S0031320321004015}
\BIBentrySTDinterwordspacing

\bibitem{bekele2019arl}
E.~Bekele and W.~Lawson, ``Multi-attribute residual network (maresnet) for
  soft-biometrics recognition in surveillance scenarios,'' in \emph{IEEE
  International Conference on Advanced Video and Signal-Based Surveillance
  (AVSS)}, 2019.

\bibitem{Rey_2025}
L.~Rey, A.~M. Bernardos, A.~D. Dobrzycki, D.~Carrami\~{n}ana, L.~Bergesio,
  J.~A. Besada, and J.~R. Casar, ``A performance analysis of you only look once
  models for deployment on constrained computational edge devices in drone
  applications,'' \emph{Electronics}, vol.~14, no.~3, p. 638, 2025.

\bibitem{nvidia_pytorchquant}
{NVIDIA Corporation}, ``Tensorrt pytorch-quantization toolkit,''
  \url{https://github.com/NVIDIA/TensorRT/tree/main/tools/pytorch-quantization},
  2024, quantConv2d, TensorQuantizer, quantized model zoo.

\bibitem{yolo11_ultralytics}
G.~Jocher and J.~Qiu, ``Ultralytics {YOLO11},''
  \url{https://github.com/ultralytics/ultralytics}, 2024.

\bibitem{he2016resnet}
K.~He, X.~Zhang, S.~Ren, and J.~Sun, ``Deep residual learning for image
  recognition,'' in \emph{Proceedings of the IEEE Conference on Computer Vision
  and Pattern Recognition (CVPR)}, 2016, pp. 770--778.

\bibitem{cuda_programming_guide}
{NVIDIA Corporation}, ``Cuda c++ programming guide: Concurrent kernel
  execution,''
  \url{https://docs.nvidia.com/cuda/cuda-c-programming-guide/index.html#concurrent-kernel-execution},
  2024, section on multi-stream concurrency and SM resource contention.

\bibitem{nvidia_dla_perf}
------, ``Maximizing deep learning performance on {NVIDIA} {Jetson} {Orin} with
  {DLA},''
  \url{https://developer.nvidia.com/blog/maximizing-deep-learning-performance-on-nvidia-jetson-orin-with-dla/},
  2023, dLA performance per watt is 3--5$\times$ compared to GPU.

\bibitem{huang2017densenet}
G.~Huang, Z.~Liu, L.~van~der Maaten, and K.~Q. Weinberger, ``Densely connected
  convolutional networks,'' in \emph{Proceedings of the IEEE Conference on
  Computer Vision and Pattern Recognition (CVPR)}, 2017, pp. 4700--4708.

\bibitem{sandler2018mobilenetv2}
M.~Sandler, A.~Howard, M.~Zhu, A.~Zhmoginov, and L.-C. Chen, ``{MobileNetV2}:
  Inverted residuals and linear bottlenecks,'' in \emph{Proceedings of the IEEE
  Conference on Computer Vision and Pattern Recognition (CVPR)}, 2018, pp.
  4510--4520.

\bibitem{howard2019mobilenetv3}
A.~Howard, M.~Sandler, G.~Chu, L.-C. Chen, B.~Chen, M.~Tan, W.~Wang, Y.~Zhu,
  R.~Pang, V.~Vasudevan, Q.~V. Le, and H.~Adam, ``Searching for
  {MobileNetV3},'' in \emph{Proceedings of the IEEE/CVF International
  Conference on Computer Vision (ICCV)}, 2019, pp. 1314--1324.

\bibitem{tan2019efficientnet}
\BIBentryALTinterwordspacing
M.~Tan and Q.~V. Le, ``{EfficientNet}: Rethinking model scaling for
  convolutional neural networks,'' in \emph{Proceedings of the International
  Conference on Machine Learning (ICML)}, 2019, pp. 6105--6114. [Online].
  Available: \url{https://proceedings.mlr.press/v97/tan19a.html}
\BIBentrySTDinterwordspacing

\end{thebibliography}

\end{document}